\documentclass[10pt,twocolumn,letterpaper]{article}

\usepackage{wacv}
\usepackage{times}
\usepackage{epsfig}
\usepackage{graphicx}
\usepackage{amsmath}
\usepackage{amssymb}

\def\wacvPaperID{951} 

\wacvfinalcopy 

\ifwacvfinal
\fi

\usepackage[breaklinks=true,bookmarks=false]{hyperref}

\usepackage{bbm}
\usepackage{subfiles}
\usepackage{booktabs}
\usepackage{tabularx}
\usepackage{arydshln}
\usepackage{xspace}
\usepackage[dvipsnames]{xcolor}
\usepackage[tight]{subfigure}
\usepackage{flushend}
\usepackage{multirow}
\usepackage{bm}
\usepackage[T1]{fontenc}
\usepackage[export]{adjustbox}
\usepackage{enumitem}
\setitemize{noitemsep,topsep=0pt,parsep=0pt,partopsep=0pt}

\usepackage[numbers]{natbib}
\bibpunct{[}{]}{,}{n}{}{;}

\begin{document}

\title{Boosting Contrastive Self-Supervised Learning with False Negative Cancellation}

\author{First Author\\
Institution1\\
Institution1 address\\
{\tt\small firstauthor@i1.org}
\and
Second Author\\
Institution2\\
First line of institution2 address\\
{\tt\small secondauthor@i2.org}
}

\author{\normalsize Tri Huynh\thanks{Work done during an internship at Google. Correspondence to Tri Huynh <trihuynh@google.com>, Maryam Khademi <maryamkhademi@google.com>.} \\ \small Google \and \normalsize Simon Kornblith \\ \small Google Research \and \normalsize Matthew R.\ Walter \\ \small TTI-Chicago \and \normalsize Michael Maire \\ \small University of Chicago \and \normalsize Maryam Khademi \\ \small Google}

\maketitle

\ifwacvfinal
\thispagestyle{empty}
\fi

\begin{abstract}
Self-supervised representation learning has made significant leaps fueled by progress in contrastive learning, which seeks to learn transformations that embed positive input pairs nearby, while pushing negative pairs far apart. 
While positive pairs can be generated reliably (e.g., as different views of the same image), it is difficult to accurately establish negative pairs, defined as samples from different images regardless of their semantic content or visual features. A fundamental problem in contrastive learning is mitigating the effects of false negatives. Contrasting false negatives induces two critical issues in representation learning: discarding semantic information and slow convergence. 
In this paper, we propose novel approaches to identify false negatives, as well as two strategies to mitigate their effect, i.e. false negative elimination and attraction,
while systematically performing rigorous evaluations to study this problem in detail. Our method exhibits consistent improvements over existing contrastive learning-based methods. Without labels, we identify false negatives with $\sim$$40\%$ accuracy among $1000$ semantic classes on ImageNet, and achieve $5.8\%$ absolute improvement in top-1 accuracy over the previous state-of-the-art when finetuning with $1\%$ labels. Our code is available at \href{https://github.com/google-research/fnc}{\color{red}{https://github.com/google-research/fnc}}
\end{abstract}

\vspace{-15pt}
\section{Introduction}
Representation learning has become the backbone of most modern AI agents. High quality pretrained representations are essential to improving performance on downstream tasks~\cite{pmlr-v32-donahue14, 6909475, 10.1007/978-3-319-10590-1_53,kim17}. While conventional approaches rely on labeled data, there has been a recent surge in self-supervised representation learning~\cite{gidaris2018unsupervised, doersch2016unsupervised, noroozi2017unsupervised, pascal2018extracting, pathak2016context, chen2020generative, 8099559, 8099579}. In fact, self-supervised representation learning has been closing the gap with and, in some cases, even surpassing its supervised counterpart~\cite{chen2020simple, he2019moco, chen2020mocov2, chen2020big}. Notably, most state-of-the-art methods are converging around and fueled by the central concept of contrastive learning~\cite{oord2019representation, henaff2020dataefficient, hjelm2018learning, tian2020contrastive, misra2019selfsupervised, he2019moco, chen2020simple}.

\begin{figure}[t!]
    \centering
    \includegraphics[width=1.0\linewidth]{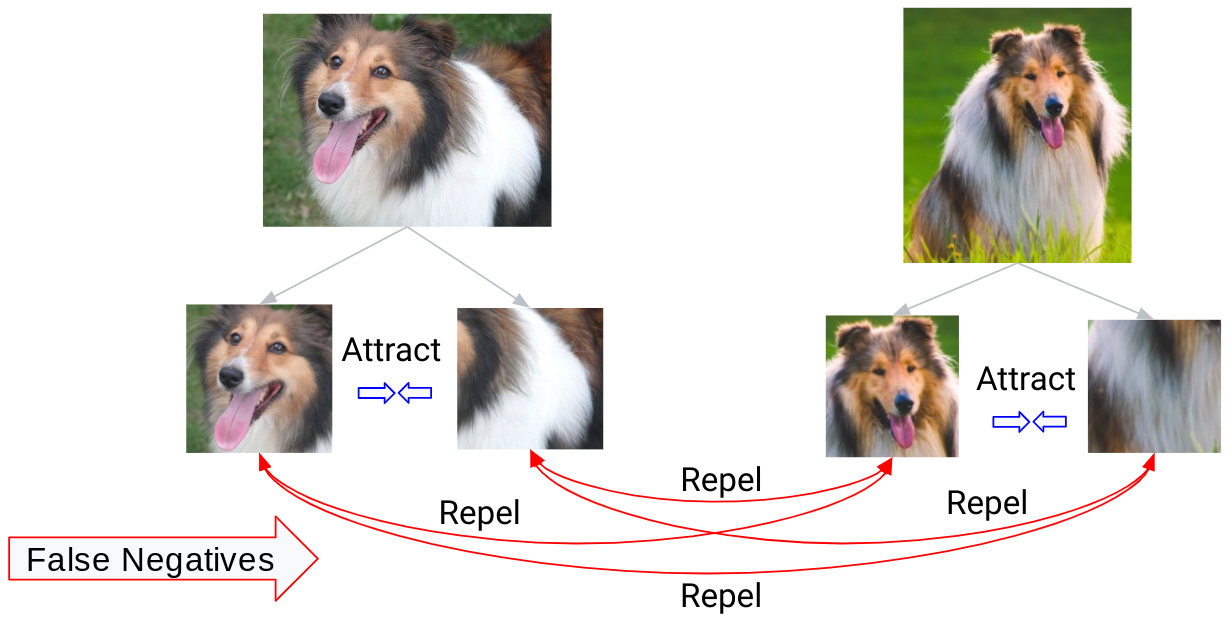}
    \caption{False negatives in contrastive learning. Without knowledge of labels, automatically selected negative pairs could actually belong to the same semantic category, creating false negatives.}
    \label{fig:false_negatives}
    \vspace{-10pt}
\end{figure}
\begin{figure*}[t!]
    \centering
    \includegraphics[width=1.0\linewidth]{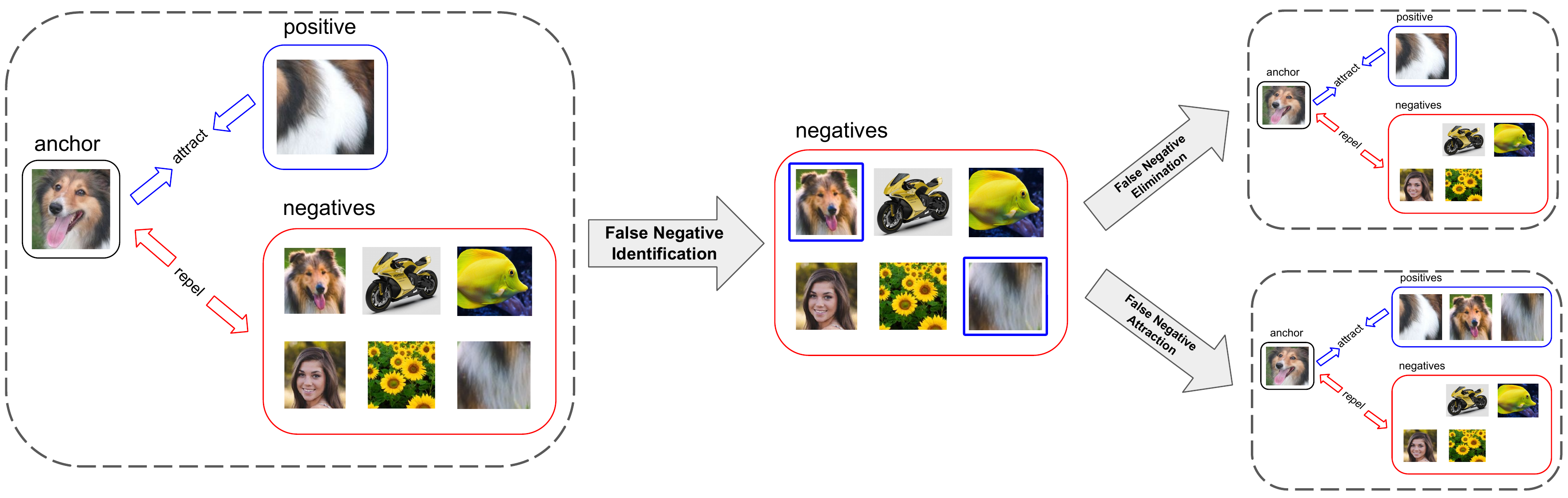}
    \caption{Overview of the proposed framework. \textbf{Left:} Original definition of the anchor, positive, and negative samples in contrastive learning. \textbf{Middle:} Identification of false negatives (blue). \textbf{Right:} false negative cancellation strategies, \emph{i.e.}~elimination and attraction.}
    \label{fig:overview}
    \vspace{-5pt}
\end{figure*}

In contrastive learning, the embedding space is governed by two opposing forces, the attraction of positive pairs and repellence of negative pairs, effectively actualized through the contrastive loss. Without labels, recent breakthroughs rely on the instance discrimination task in which positive pairs are defined as different views of the same image, while negative pairs are formed by sampling views from different images, regardless of their semantic information~\cite{he2019moco, chen2020simple, misra2019selfsupervised}. Figure~\ref{fig:false_negatives} illustrates this process. Positive pairs generated from different views of the same image are generally reliable since they likely contain similar semantic content or correlated features~\cite{geirhos2020shortcut}. However, the creation of valid negative pairs is far more difficult. The common approach of defining negative pairs as samples from different images ignores their semantic content.
For example, two images of a dog are considered a negative pair, as seen in Figure~\ref{fig:false_negatives}.

Contrasting undesirable negative pairs encourages the model to discard their common features through the embedding, which are indeed the common semantic content, \emph{e.g.}, dog features in the previous example. We define those undesirable negatives as false negatives, \emph{i.e.}, negative pairs from the same semantic category. Besides disregarding semantic information, false negatives also hinder the convergence of contrastive learning-based objectives due to the appearance of contradicting objectives. For instance, in Figure~\ref{fig:false_negatives}, the dog's head on the left is attracted to its fur (positive pair), but repelled from similar fur of another dog image on the right (negative pair), creating contradicting objectives.

While recent efforts focus on improved architectures~\cite{chen2020simple, chen2020big, he2019moco} and data augmentation~\cite{chen2020simple, tian2020what}, relatively little work considers the effects of negative samples, especially that of false negatives. Most existing methods focus on mining hard negatives~\cite{robinson2020contrastive, kalantidis2020hard}, or most recently, reweighing positive and negative terms~\cite{chuang2020debiased}. However, there has been little effort attempting to identify false negatives.

False negatives remain a fundamental problem in contrastive self-supervised learning. Without labels, this problem is very hard to adequately resolve, as it boils down to a chicken-and-egg problem, where we want to learn good semantic representations, but may need certain semantic information to start with. Nevertheless, in this paper, we attempt to study this problem in detail and propose novel ideas to overcome its limitations, as overviewed in Figure \ref{fig:overview}. Particularly, the contributions of the paper are as follows:
\begin{itemize}
    \item We propose simple yet effective strategies to find potential false negatives in contrastive learning. Without  labels, the method effectively finds false negatives with $\sim$40\% accuracy among $1000$ semantic categories on ImageNet~\cite{imagenet}.
    \item We propose and study the effect of two different strategies to improve the contrastive loss based on the estimated false negatives, \emph{i.e.}, false negative elimination and attraction. The experiments reveal the significant impact false negatives have on contrastive learning.
    \item Our method consistently improves over existing approaches across a wide range of settings, \emph{e.g.}, with or without momentum contrast~\cite{he2019moco}. Further, we show that our methods are effectively complementary with the recent multi-crop augmentation strategy in both accuracy and efficiency: Adding false negative cancellation on top of multi-crop incurs negligible overhead, while doubling the improvement (4\% in ImageNet).
    \item The improved model establishes new state-of-the-art results for contrastive learning-based methods across all evaluations on ImageNet, as well as transferring to downstream tasks. Notably, the model achieves 5.8\% absolute improvement of Top-1 accuracy in the semi-supervised setting with 1\% labels on ImageNet.
\end{itemize}


\section{Related Work}
\label{sec:related_work}

Early work employs proxy tasks to guide the learned embeddings, \emph{e.g.} ~\cite{gidaris2018unsupervised}, ~\cite{doersch2016unsupervised}, ~\cite{noroozi2017unsupervised}, ~\cite{pascal2018extracting}. While effective, the proxy tasks are heuristic and lack generality.

Other works use clustering-based methods~\cite{asano2020self, cliquecnn2016, 10.5555/3045390.3045442}. \citet{caron2018deep} iteratively improve the representations by clustering samples and using these clusters as pseudo-labels. They then train the network to classify samples based on these pseudo-labels. While our elimination-based strategy is distinct from clustering, our attraction-based approach also attempts to group visually connected samples, yet, the formulation and context differ.

We take a contrastive learning-based approach to self-supervised representation learning. 
Earlier work in this area includes CPC~\cite{oord2019representation, henaff2020dataefficient}, Deep InfoMax~\cite{hjelm2018learning,bachman2019learning}, and CMC~\cite{tian2020contrastive}. Recognizing the importance of a large negative pool, PIRL~\cite{misra2019selfsupervised} maintains a memory bank of all image representations, which limits scalability. MoCo~\cite{he2019moco} addresses this using a momentum encoder. SimCLR~\cite{chen2020simple} eschews a momentum encoder in favor of a large batch size, and proposes updates to the projection head and data augmentation.

Realizing the important role of negative samples in contrastive learning, a few recent methods investigate ways to improve negative sampling. However, most works focus on the effect of hard negatives (\emph{i.e.}, the true negatives that are close to the anchor, which are distinct from false negatives)~\cite{robinson2020contrastive, kalantidis2020hard}. Arguing that hard negatives could be noisy, Ring~\cite{wu2021conditional} proposes to sample negatives within percentile ranges. Besides hard negatives, other methods extend to also consider hard positives. While ~\cite{wang2020advances} performs hard selection of samples, ~\cite{zheng2021contrastive} utilizes soft selection by conditional probability.
~\citet{chuang2020debiased} reweigh positive and negative terms to reduce the effects of undesirable negatives. While formulated differently, their approach is similar to multi-crop~\cite{caron2020unsupervised}, in that it increases the contribution of positive terms by subtracting them in the denominator of the contrastive loss, whereas multi-crop does so by adding positive terms in the numerator. However, both methods fail to identify false negatives and lack semantic feature diversity. 



%

Inspired by contrastive learning, recent methods predict one positive view from another. However, these methods are not categorized as contrastive learning. They have different formulations at their core and do not contrast against negative samples, a defining element of contrastive learning. SwAV~\cite{caron2020unsupervised} is an online clustering-based method that employs swapping prediction of different views from an image, while BYOL~\cite{grill2020bootstrap} employs a momentum encoder as a prediction target for another view from the main encoder.

\section{Method}
\label{sec:method}

\subsection{Contrastive Learning}
Contrastive learning seeks to learn a transformation that brings positive pairs ``nearby'' in an embedding space while pushing negative pairs apart. This is done by minimizing a contrastive loss that, for each \emph{anchor} image $i$, measures the (negative) similarity between its embedding $z_i$ and that of its positive match $z_j$ relative to the similarity between the anchor embedding of $k \in \{1, \ldots, M\}$ negative matches:
%
%
\begin{equation}
    l_{i}=-\log\frac{\exp(\textrm{sim}(z_i,z_j)/\tau)}{\sum_{k=1}^{M} \mathbbm{1}_{[k\ne i]}\exp(\textrm{sim}(z_i,z_k)/\tau)},
\label{eq:contrastive_loss}
\end{equation}
where $\textrm{sim}(u,v)$ is a similarity function, \emph{e.g.}, the $L_2$ normalized cosine similarity $\textrm{sim}(u,v) = u^Tv/\lVert u\rVert \lVert v\rVert$, and $\tau$ is a temperature parameter.

Without known correspondence, self-supervised methods commonly define positive pairs as different augmentations of the same image and negative pairs as samples from different images. Consider a batch of $N$ images, each augmented to form $N$ positive pairs for a total of $2N$ images. Methods like SimCLR use samples from the same batch as negative pairs (\emph{i.e.}, $M = 2N$ examples for each anchor $i$), while MoCo uses samples from a momentum encoder to avoid the use of a large batch size.
%
%
%
%


\subsection{False Negative Cancellation}
Negative pairs are samples from different images that may or may not possess similar semantic content or visual features. Consequently, it is possible that some samples $k$ have the same semantic content as the anchor $i$, and are thus false negatives. As discussed earlier, false negatives give rise to two critical problems in contrastive learning: they discard semantic information and slow convergence.


Supposing we can find the false negatives, we propose two strategies that use them to improve contrastive learning.

\subsubsection{False Negative Elimination}
The simplest strategy for mitigating the effects of false negatives is to not contrast against them. This amounts to the following modification to the contrastive objective~\eqref{eq:contrastive_loss}:
\begin{equation} \label{eq:elimination_loss}
l_{i}^\textrm{elim}=-\log\frac{\exp(\textrm{sim}(z_i,z_j)/\tau)}{\sum_{k=1}^{2N} \mathbbm{1}_{[k\ne i, k \notin \mathbbm{F}_i]}\exp(\textrm{sim}(z_i,z_k)/\tau)},
\end{equation}
where $\mathbbm{F}_i$ is the set of the detected false negatives with respect to an anchor $i$.

\subsubsection{False Negative Attraction}
While eliminating false negatives alleviates the undesireable effects of contrasting against them, it ignores information available in what are actually true positives. Minimizing the original contrastive loss (Eqn.~\ref{eq:contrastive_loss}) only seeks to attract an anchor to different views of the same image. Including true positives drawn from different images would increase the diversity of the training data and, in turn, has the potential to improve the quality of the learned embeddings. Indeed, \citet{khosla2020supervised} show that supervised contrastive learning (\emph{i.e.}, where an anchor is attracted to samples having the same semantic label) can be more effective than the traditional supervised cross-entropy loss. Thus, we propose to treat the false negatives that have been identified as true positives and attract the anchor to this set. This yields the following expression for the contrastive loss:
\begin{multline}
l_{i}^\textrm{att}= - \frac{1}{1+\lvert \mathbbm{F}_i \rvert} \left( \log\frac{\exp(\textrm{sim}(z_i,z_j)/\tau)}{\sum_{k=1}^{2N} \mathbbm{1}_{[k\ne i]}\exp(\textrm{sim}(z_i,z_k)/\tau)} \right.\\
               + \sum_{f\in \mathbbm{F}_i} \left. \log\frac{\exp(\textrm{sim}(z_i,z_f)/\tau)}{\sum_{k=1}^{2N} \mathbbm{1}_{[k\ne i]}\exp(\textrm{sim}(z_i,z_k)/\tau)} \right)
\end{multline}
%
%


\subsubsection{Finding False Negatives} \label{sec:find_fn}

\begin{figure}[t]
\begin{center}
    \includegraphics[width=1.0\linewidth]{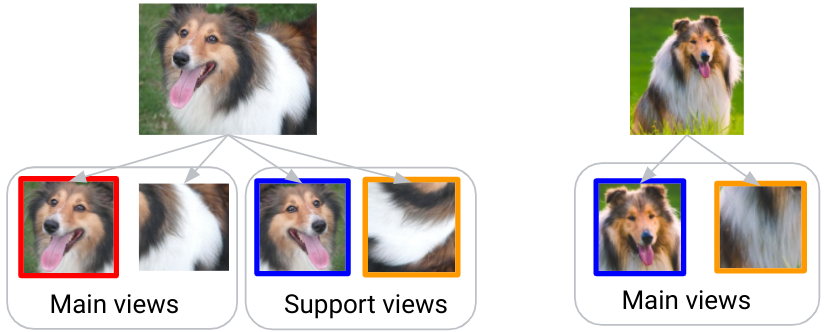}
    \caption{Finding false negatives with the support views. Negative samples (main views, right) may not have as reliable similarity with the anchor itself (red) as they do with other augmented views of the same image (support views). For instance, the dog's face in the support view (left, blue) is closer to the negative sample (right, blue) in terms of the facial orientation than the anchor (red).}\label{fig:find_fn}
\end{center}
\vspace{-10pt}
\end{figure}

Unfortunately, the process of identifying false negatives is fundamentally difficult, amounting to a chicken-and-egg problem---without labels, the learned semantic information can be used to establish valid and invalid correspondences, yet the correctness of these embeddings depends on the ability to identify correspondences. Our approach to identify false negatives based on the following observations:
\textit{
\begin{itemize}
    \vspace{5pt}
    \item False negatives are samples from different images with the same semantic content, therefore they should hold certain similarity (\emph{e.g.}, dog features).
    \item A false negative may not be as similar to the anchor as it is to other augmentations of the same image, as each augmentation only holds a specific view of the object (Figure~\ref{fig:find_fn}).
    \vspace{5pt}
\end{itemize}
}

The above observations mean that we may be able to approximate a false negative with more augmented views of the anchor. As an example, consider Figure~\ref{fig:find_fn} where we treat the picture of the dog's head on the left (``main views,'' in red) as the anchor image. The support views on the left are other augmented views generated from the same image and serve as positive matches. The picture of the dog's head on the far right (``main view,'' in blue) is not an augmented version of the anchor. Consequently, while it is similar to the anchor image, it would thus be treated as a negative match by contemporary self-supervised methods (\emph{i.e.}, it is a false negative). However, we see that this image is more similar to the augmented view of the anchor (``support view,'' in blue) than it is to the anchor with respect to the orientation of the dog's face. Similarly, the head and fur of the dog are expected to be a positive pair. While the false negative fur on the far right (orange) could look different than the head anchor, it should be more similar to the fur in the support views (orange).

Motivated by these observations, we propose a strategy for identifying candidate false negatives that follows as:
\begin{enumerate}
    \item For each anchor $i$, generate a support set $\mathbbm{S}_i=\{z_i^s\}$ that contains other support views from the same image besides the two main views.
    \item Compute similarity scores, $\textrm{score}_{m,i}^s=\textrm{sim}(z_m,z_i^s)$, between a negative sample $z_m$
    and each sample $z_i^s$ in the support set.
    \item Aggregate the computed scores for each negative sample, $\textrm{score}_{m,i}=\textrm{aggregate}_{s\in \mathbbm{S}}(\textrm{score}_{m,i}^s)$.
    \item Define a set of potential false negatives $\mathbbm{F}_i$ as the negative samples that are most similar to the support set based on the aggregated scores, $\mathbbm{F}_i=\textrm{best}(\textrm{score}_{i})$, where $\textrm{score}_i=\{\textrm{score}_{m,i} \vert m\}$ is the set of scores for each negative sample with respect to anchor $i$.
\end{enumerate}
For each element in the above procedure, there are several considerations one can make, including the choice of the similarity function, the strategy for aggregating scores, and the manner in which the most similar samples are defined. In this work, we investigate the following options:
\setlist[description]{leftmargin=0pt,labelindent=0pt}
\begin{description}
    \item[Similarity Function] We use the cosine similarity function, since it is used in the contrastive loss during pretraining.
    \item[Aggregation Strategy] We consider both mean aggregation, $\textrm{score}_{m,i}=\frac{1}{\lvert \mathbbm{S} \rvert}\sum_{s=1}^{\lvert \mathbbm{S} \rvert}\textrm{sim}(z_m,z_i^s)$, and max aggregation, $\textrm{score}_{m,i}=\max_{s \in \{1, \ldots, \lvert \mathbbm{S} \rvert\}}\textrm{sim}(z_m,z_i^s)$, and discuss their effects in Section~\ref{sec:experiments}.
    \item[Screening Strategy] We consider two choices for the most similar samples, one that considers the top-$k$ matches, $\textrm{best}(\textrm{score}_{i})=\{z_m \vert \textrm{score}_{m,i}\in \textrm{top}(\textrm{score}_{i},k)\}$, and one that considers those above a threshold $t$, $\textrm{best}(\textrm{score}_{i})=\{z_m \vert \textrm{score}_{m,i}>t\}$. A top-$k$ strategy may be preferred given information about the approximate number of false negatives, while thresholding may be better suited when a dynamic adaptation is expected. We also consider a strategy that combines top-$k$ and and thresholding, $\textrm{best}(\textrm{score}_{i})=\{z_m \vert \textrm{score}_{m,i}\in \textrm{top}(\textrm{score}_{i}, k) \: \& \: \textrm{score}_{m,i}>t\}$.
\end{description}

\section{Experimental Results}
\label{sec:experiments}


\subsection{Ablation Studies}

We use the same configuration as SimCLR v2 for pretraining and evaluation. The base encoder is ResNet-50 with a 3-layer MLP projection head. Data augmentation includes random crops, color distortion, and Gaussian blur. For each experiment, we pretrain for 100 epochs on the ImageNet ILSVRC-2012 training set, then freeze the encoder and train a linear classifier on top, which is then evaluated on the ImageNet evaluation set. We pretrain on 128 Cloud TPUs with a batch size of 4096. We use the LARS optimizer~\cite{ginsburg2018large} with a learning rate of 6.4 and a cosine decay schedule, and a weight decay of $1 \times 10^{-4}$.

\subsubsection{False Negative Cancellation Strategies}

\begin{figure}[!t]
    \centering
    \includegraphics[width=0.485\linewidth]{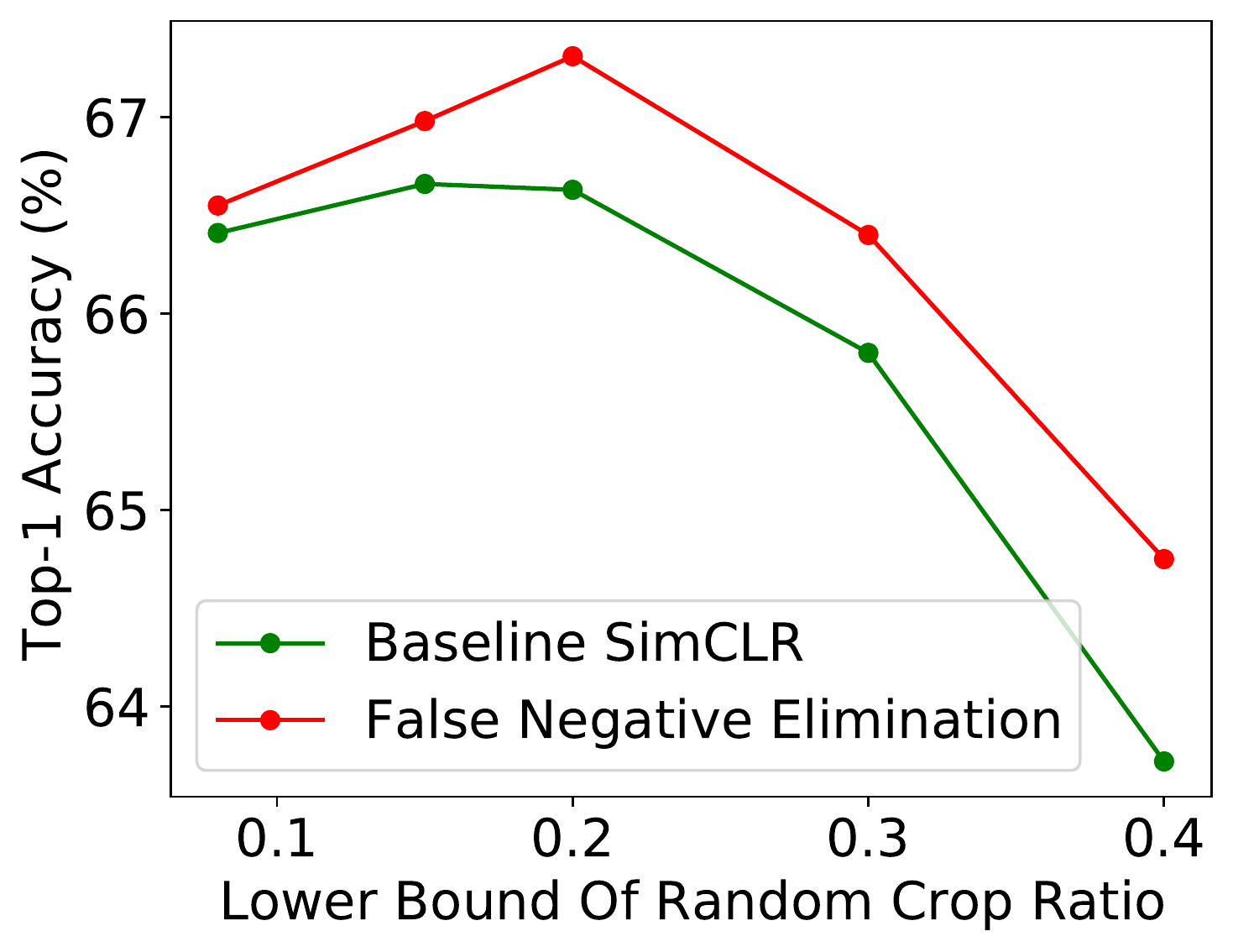} \hfil
    \includegraphics[width=0.485\linewidth]{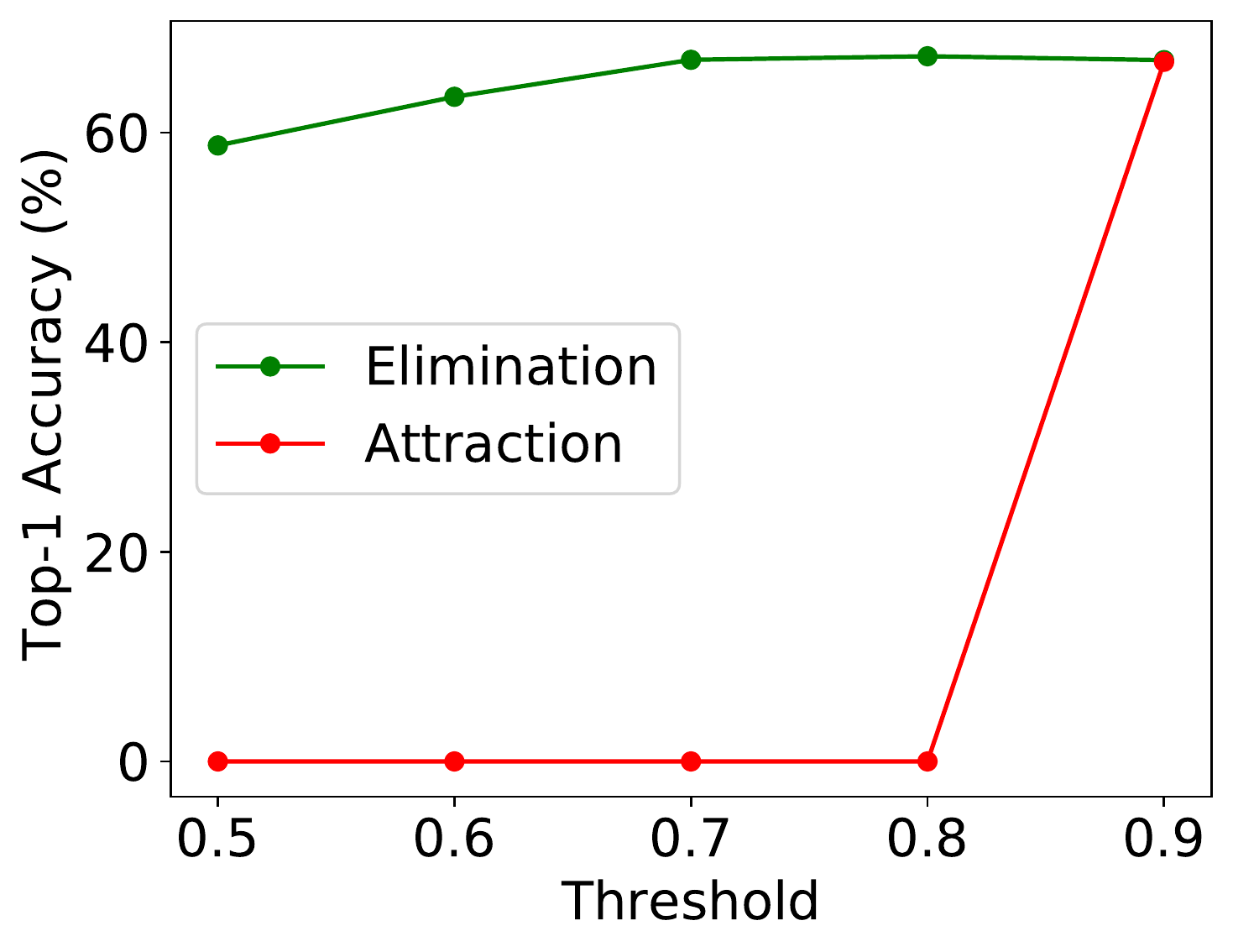}
    \caption{A comparison of top-1 accuracy (left) between false negative elimination and SimCLR (\emph{lower bound of random crop ratio} represents the lowest cropping ratio in random image augmentation); and (right) top-1 accuracy across filtering thresholds in false negative cancellation.}\label{fig:top1_accuracy}
    \vspace{-5pt}
\end{figure}
We evaluate the effects of various approaches to false negative mitigation, including the choice of cancellation strategy, aggregation score, and screening strategies, and draw the following conclusions.

\textbf{False negative elimination consistently improves contrastive learning across crop sizes, and the gap is higher for bigger crops.} Figure~\ref{fig:top1_accuracy} (left) demonstrates that the inclusion of false negative elimination yields top-1 accuracy that is strictly better than that of the SimCLR baseline accross the full range of crop ratios. We postulate that the bigger gap for larger crop sizes is due to the increased chance of having common semantic content in big crops, which leads to a higher ratio of false negatives. It is also worth noting that in Figure~\ref{fig:top1_accuracy} (left), we only eliminate a negligible number of two potential false negatives among 8190 (batch size 4096) negative samples for each anchor, but it could affect top-1 accuracy by as much as $1\%$. This supports the significant effect of false negatives in contrastive learning.

\begin{figure}[!b]
    \includegraphics[width=0.485\linewidth]{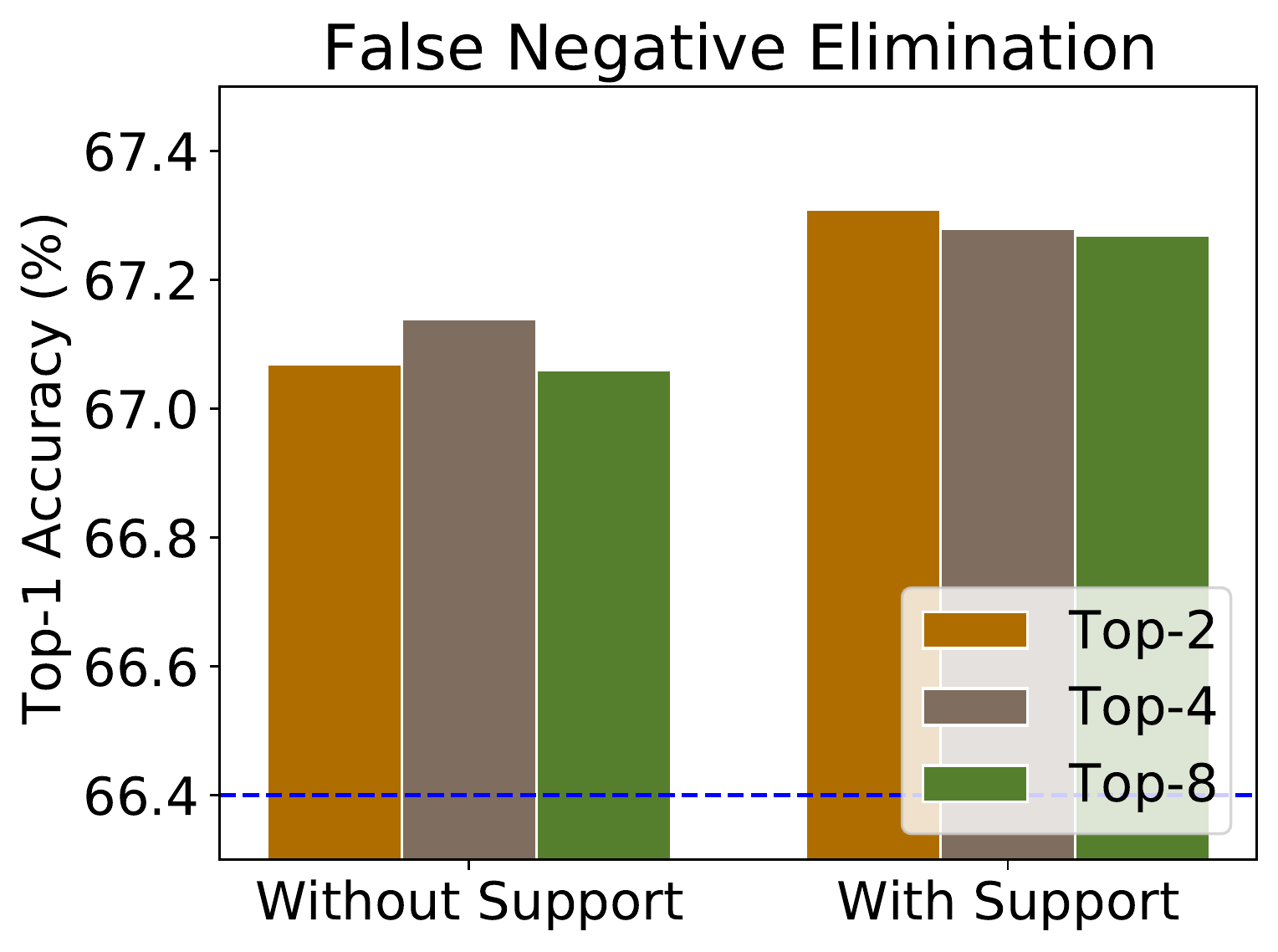} \hfil
    \includegraphics[width=0.485\linewidth]{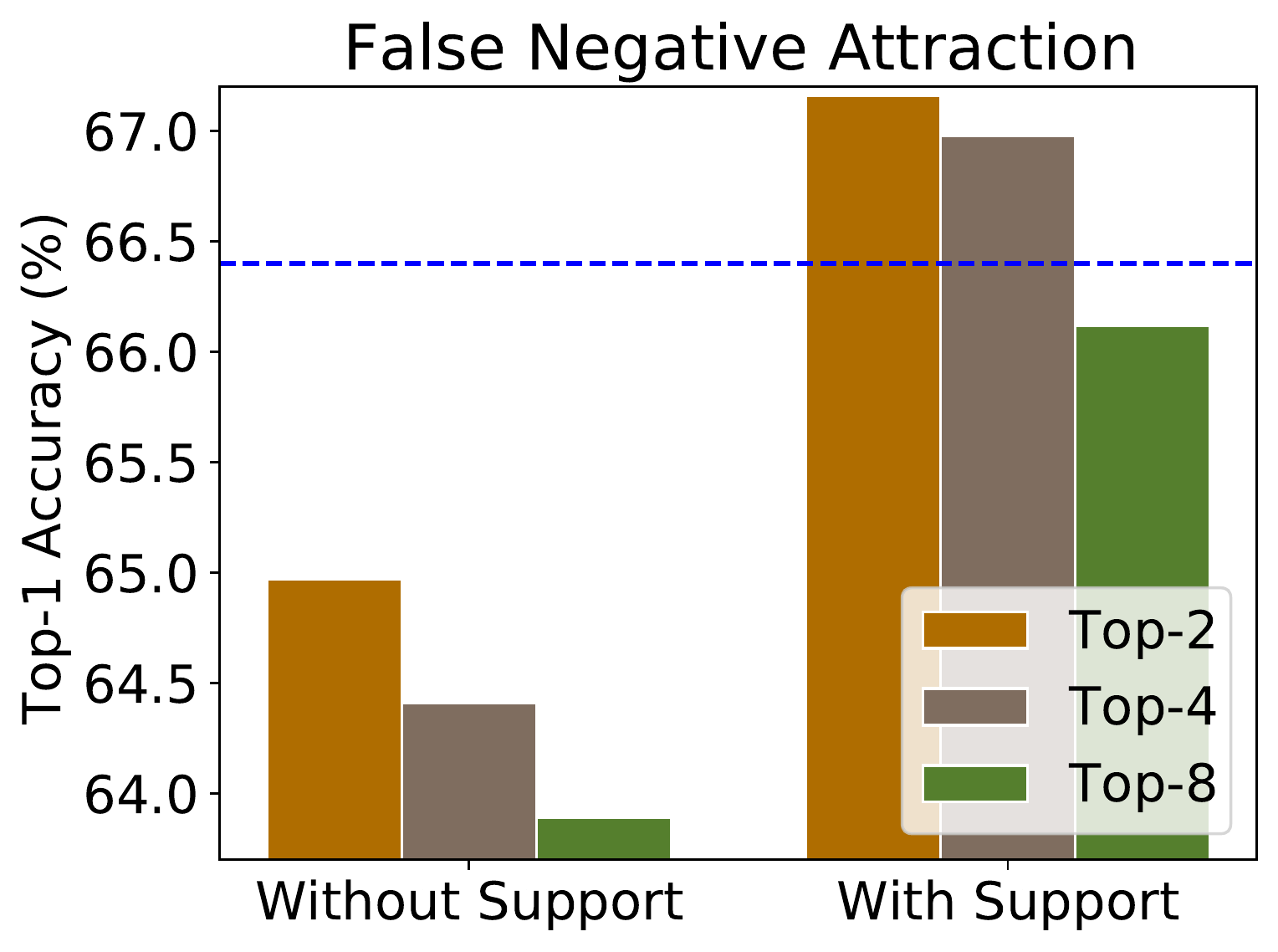}
    \caption{False negative cancellation with and without support set across top-$k$ choices for different mitigation strategies. The dashed line denotes the performance of the SimCLR baseline. The results use mean aggregation in scoring potential false negatives.}
    \label{fig:support_topk}
\end{figure}
\textbf{Having a support set helps in finding false negatives regardless of the cancellation strategy, with greater benefits with the attraction strategy.} Figure~\ref{fig:support_topk} contrasts the top-1 accuracy when we compare negative samples to a support set (Section~\ref{sec:find_fn}) to the case in which we only compare negative samples to the anchor itself. The use of a support set results in larger performance gains when using false negative attraction ($\sim$2\%) compared to the false negative elimination strategy ($\sim$0.2\%). Further, while the elimination strategy improves performance relative to the SimCLR baseline whether or not a support set is used, attracting false negatives found without support set actually hurts performance (Figure~\ref{fig:support_topk}, right). This likely results from the fact that embeddings learned with attraction strategy are more sensitive to invalid false negatives (discussed next), justifying the use of a support set to reliably find false negatives.

\textbf{The attraction strategy is much more sensitive to the quality of the found false negatives compared to the elimination strategy.} This property is consistently confirmed through Figure~\ref{fig:support_topk} (right) as previously discussed, and Figure~\ref{fig:top1_accuracy} (right), where the attraction strategy only works with more reliable false negatives, those having very high similarity scores, while the elimination method is not very sensitive to the thresholds.

\begin{figure}[!t]
    \includegraphics[width=0.485\linewidth]{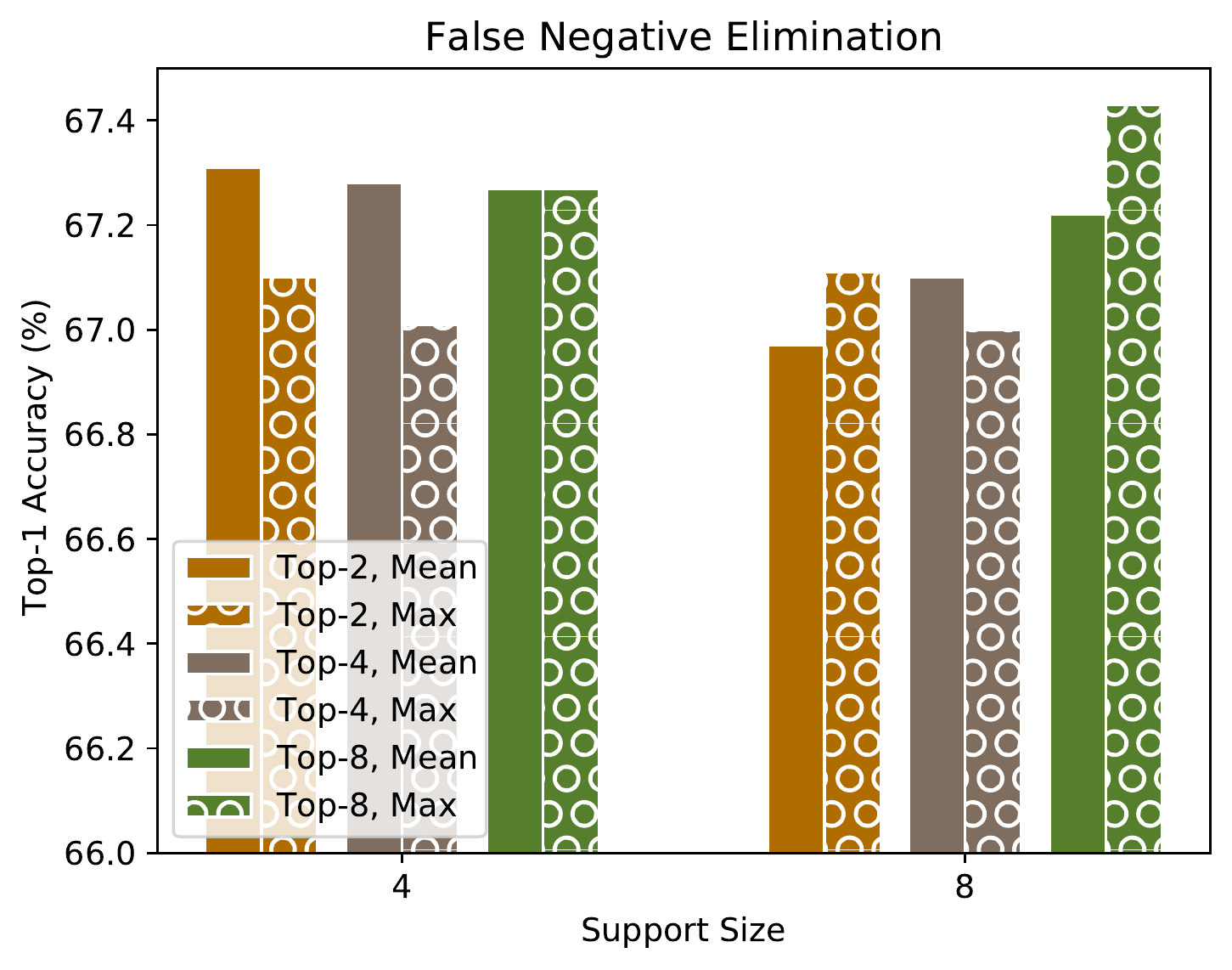} \hfil
    \includegraphics[width=0.485\linewidth]{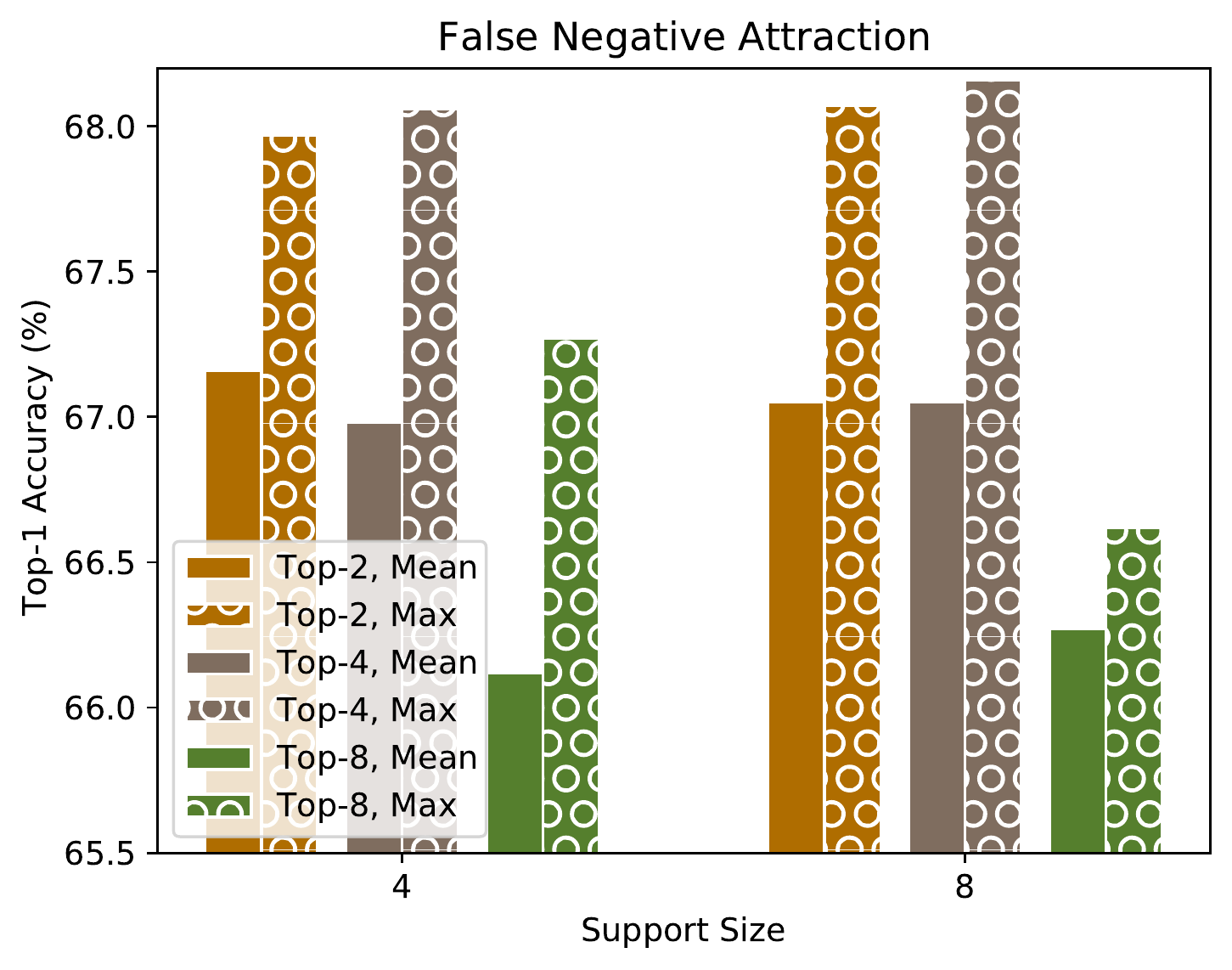}
    \caption{False negative cancellation with mean and max aggregation across support sizes and top-$k$ for the false negative (left) elimination and (right) attraction strategies.} \label{fig:mean_max}
    \vspace{-5pt}
\end{figure}
\textbf{Max aggregation significantly and consistently outperforms mean aggregation for the attraction strategy, while the gains are less pronounced with false negative elimination.} Figure~\ref{fig:mean_max} demonstrates that max aggregation outperforms mean aggregation for all support sizes and top-$k$ values in the attraction strategy, with a gap in some cases greater than $1\%$. This may be due to the fact that false negatives are similar to a strict subset of the support set, in which case considering all elements as in mean aggregation corrupts the similarity score. The difference is more pronounced for the attraction strategy, which is more sensitive to invalid false negatives.

\begin{figure}[!th]
    \includegraphics[width=0.485\linewidth,valign=t]{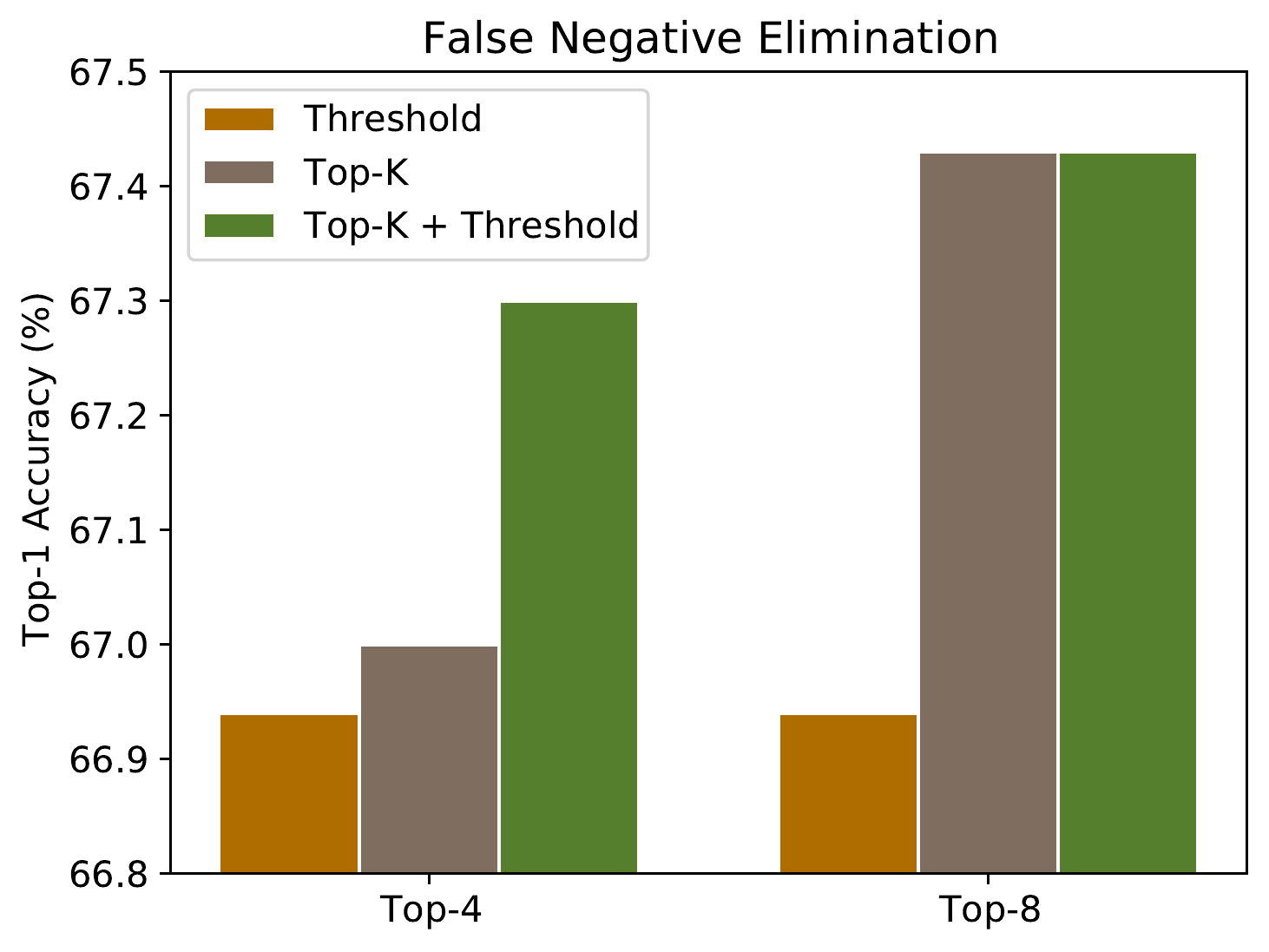}\hfil
    \includegraphics[width=0.485\linewidth,valign=t]{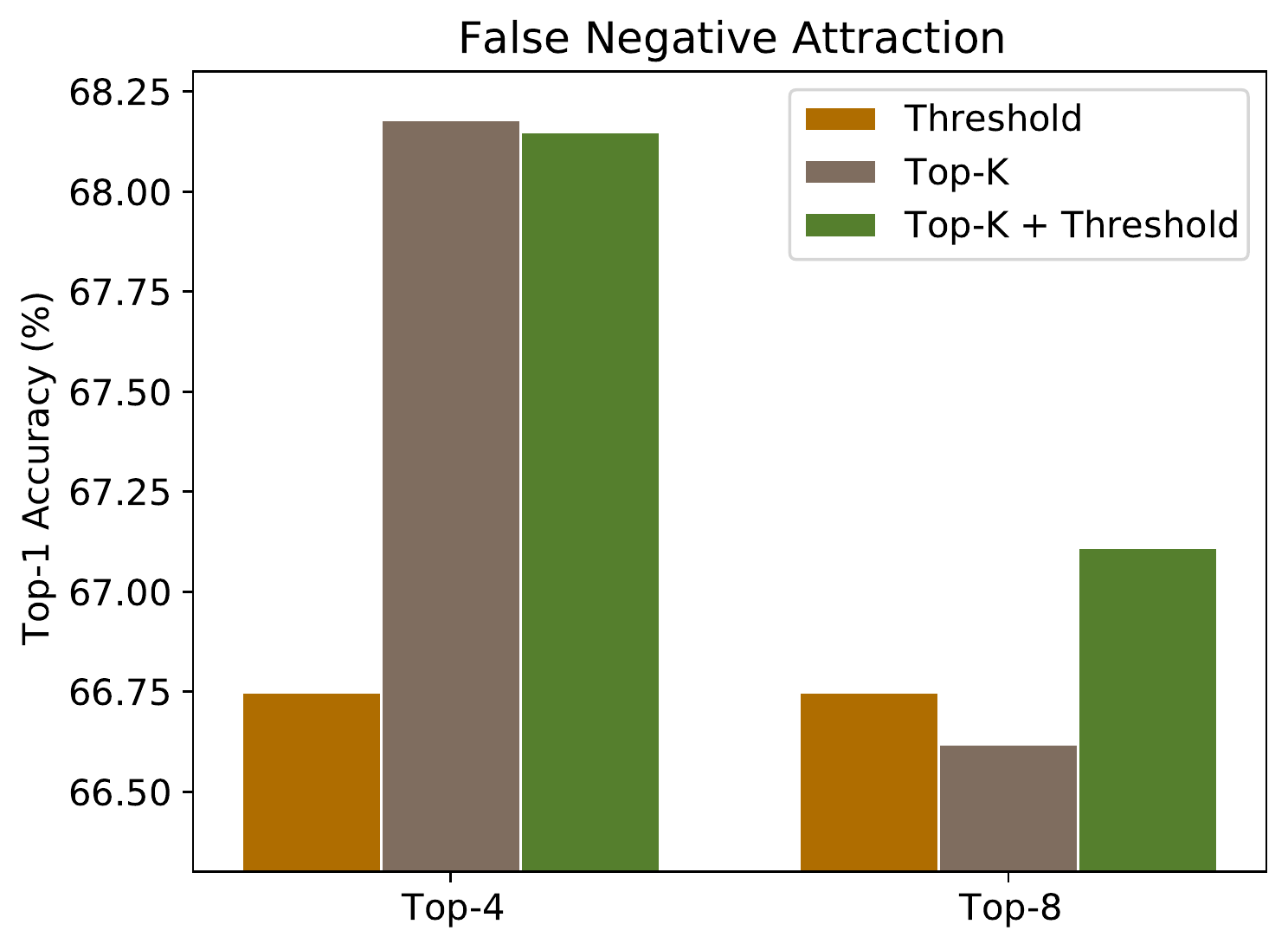}
    \caption{A comparison of top-$k$ and threshold-based filtering for false negative (left) elimination and (right) attraction strategies.} \label{fig:filtering}
\end{figure}
\textbf{Filtering by top-$\bm{k}$ tends to perform better than by a threshold, while a combination of both provides the best balance.} As seen in Figure~\ref{fig:filtering}, the best choice of top-$k$ is better than the best threshold. A strategy that combines the two approaches achieves greater accuracy, with the exception of false negative attraction at top-4, for which there is a negligible degradation in performance. 

\begin{figure}[!th]
    \includegraphics[width=0.485\linewidth,valign=t]{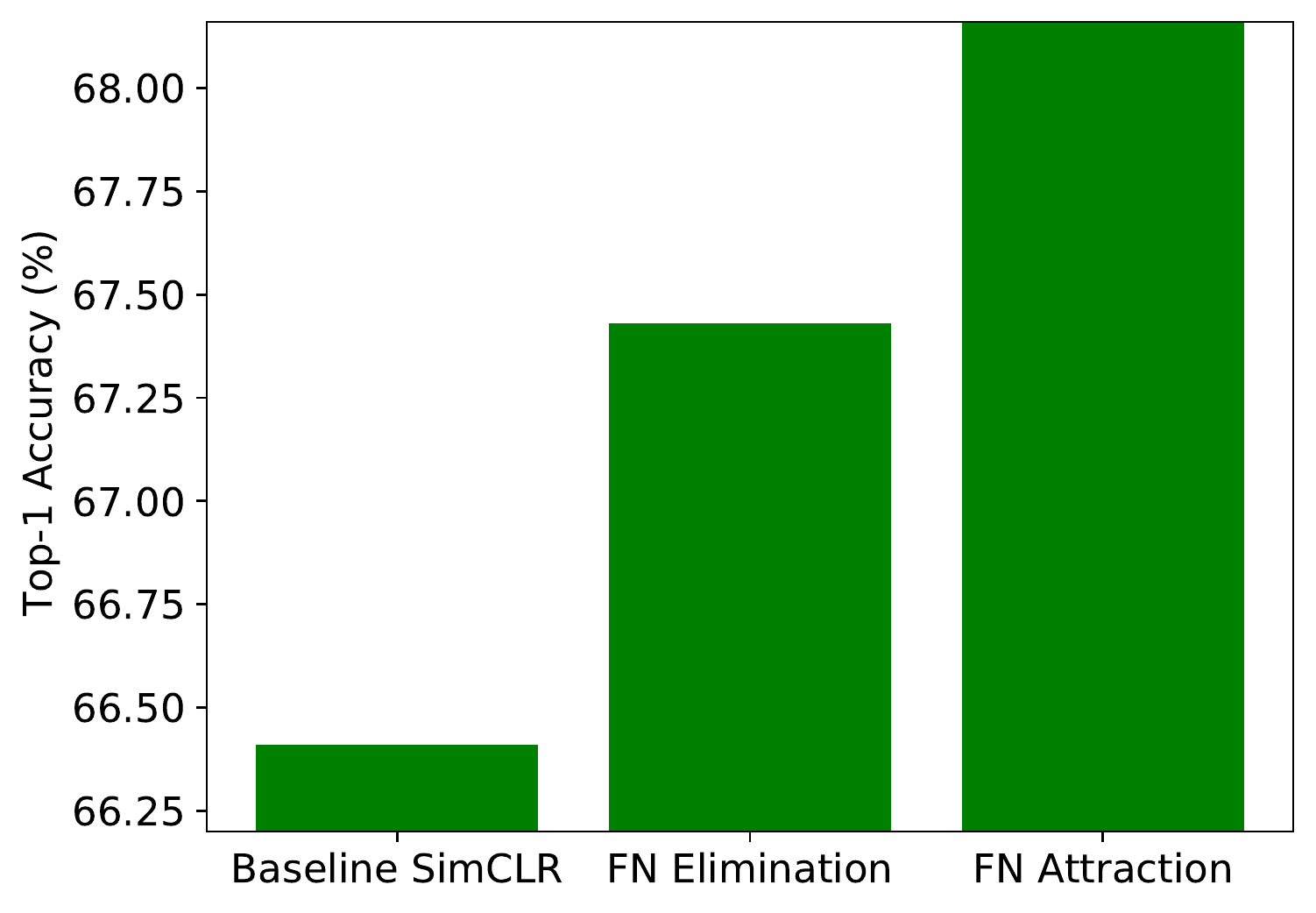}\hfil
    \includegraphics[width=0.485\linewidth,valign=t]{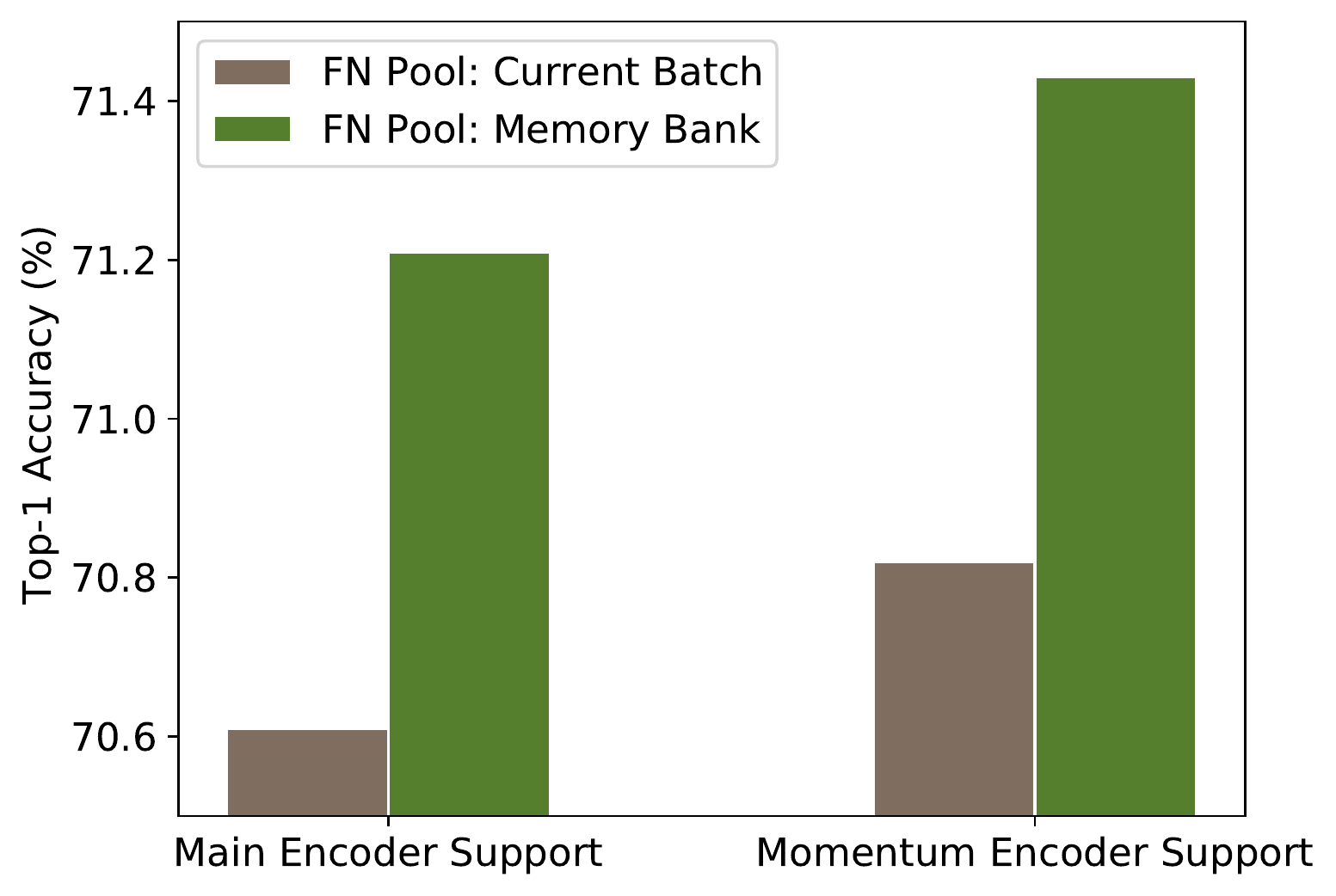}
    \caption{A comparison of top-1 accuracy (left) for different false negative cancellation strategies relative to the SimCLR baseline and (right) the use of the primary and momentum encoders.} \label{fig:baseline_vs_fn_encoder_fn_pool}
\end{figure}
\textbf{False negative attraction is superior to elimination when the detected false negatives are valid.} As shown in Figure~\ref{fig:baseline_vs_fn_encoder_fn_pool} (left) and Table~\ref{tab:fn_elim_att}, false negative elimination improves upon SimCLR by $1.02\%$, while false negative attraction results in a $1.75\%$ improvement. These results use max aggregation, a support size of eight, top-4 for false negative attraction, and top-8 filtering for elimination. As false negative attraction works better, we employ the attraction strategy in the following experiments.
\begin{table}
    \footnotesize
    \centering
    \begin{tabularx}{1.0\linewidth}{ccc}
    \toprule
    SimCLR & FN Elimination / Gain & FN Attraction / Gain\\
    \midrule
    66.41 & 67.43 / \textcolor{ForestGreen}{+1.02} & 68.16 / \textcolor{ForestGreen}{+1.75} \\
    \bottomrule
    \end{tabularx}
    \vspace{5pt}
    \caption{Top-1 accuracy improvement of false negative cancellation strategies over the SimCLR baseline.}
    \label{tab:fn_elim_att}
\end{table}




\vspace{-5pt}
\subsubsection{With Multi-crop}

\citet{caron2020unsupervised} propose a multi-crop data augmentation strategy that increases the number of positive views attracted to each anchor image, improving the quality of the learned embeddings. Multi-crop is closely related and complementary to false negative attraction in multiple facets, from principle to computational efficiency. While multi-crop attracts more positive samples, false negative attraction tries to attract samples that would otherwise be erroneously treated as negatives. The positive samples should be more reliable than the false negatives we attempt to find; however, multi-crop lacks semantic feature diversity, as it never attracts samples from different images. Because of these characteristics, multi-crop and false negative attraction offer complementary advantages. Furthermore, they can share computational overhead by using a common support set. Thus far, we have only used the support set to find false negatives, but they can also be used as additional positive views for multi-crop. In doing so, we may be able to double the performance without noticeable overhead if the respective improvements are complementary.

%
%
%
%
%
%
\begin{table}[!t]
    \centering
    {\footnotesize
    \begin{tabularx}{\linewidth}{Xr@{ }lr@{ }l}
    \toprule
    Method & Accuracy & (Diff.) & Time & (Diff.)\\
    \midrule
    SimCLR & 66.41 & & 2.63 & \\
    SimCLR + Multi-crop &  68.50 & (\textcolor{ForestGreen}{+2.09}) &  7.40 & (\textcolor{red}{+4.77})\\
    SimCLR + Multi-crop + FN Att.\ &  70.42 & (\textcolor{ForestGreen}{+1.92}) & 7.50 & (\textcolor{red}{+0.10})\\
    \bottomrule
    \end{tabularx}}
    \vspace{3pt}
    \caption{Complementary performance and computational efficiency of multi-crop and false negative attraction.} \label{tab:efficiency}
\end{table}
Figure~\ref{fig:multicrop_momentum} and Table~\ref{tab:efficiency} demonstrate the advantages of using false negative cancellation together with multi-crop data augmentation. Adding multi-crop improves the performance of the SimCLR baseline by $2.09\%$, while incurring an additional 4.77 hours of computational overhead. Further adding false negative attraction on top of multi-crop yields a $1.92\%$ absolute improvement in accuracy (i.e., similar to the $2.09\%$ gain provided by multi-crop), while incurring only $0.1$ hours of computation time.

\vspace{-5pt}
\subsubsection{With Momentum Encoders}
\begin{figure}[t]
    \centering
    \includegraphics[width=0.485\linewidth,valign=t]{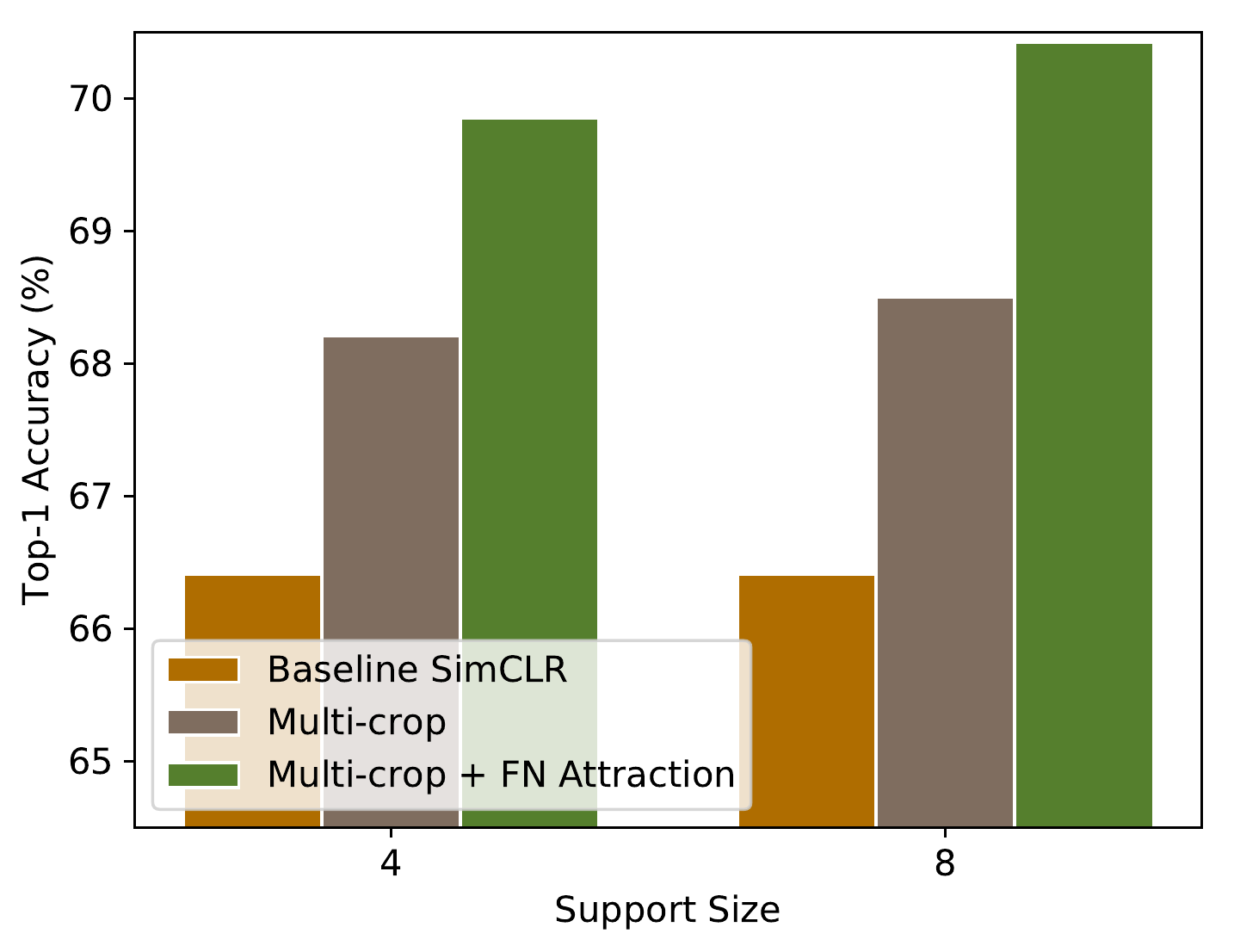}\hfil
    \includegraphics[width=0.485\linewidth,valign=t]{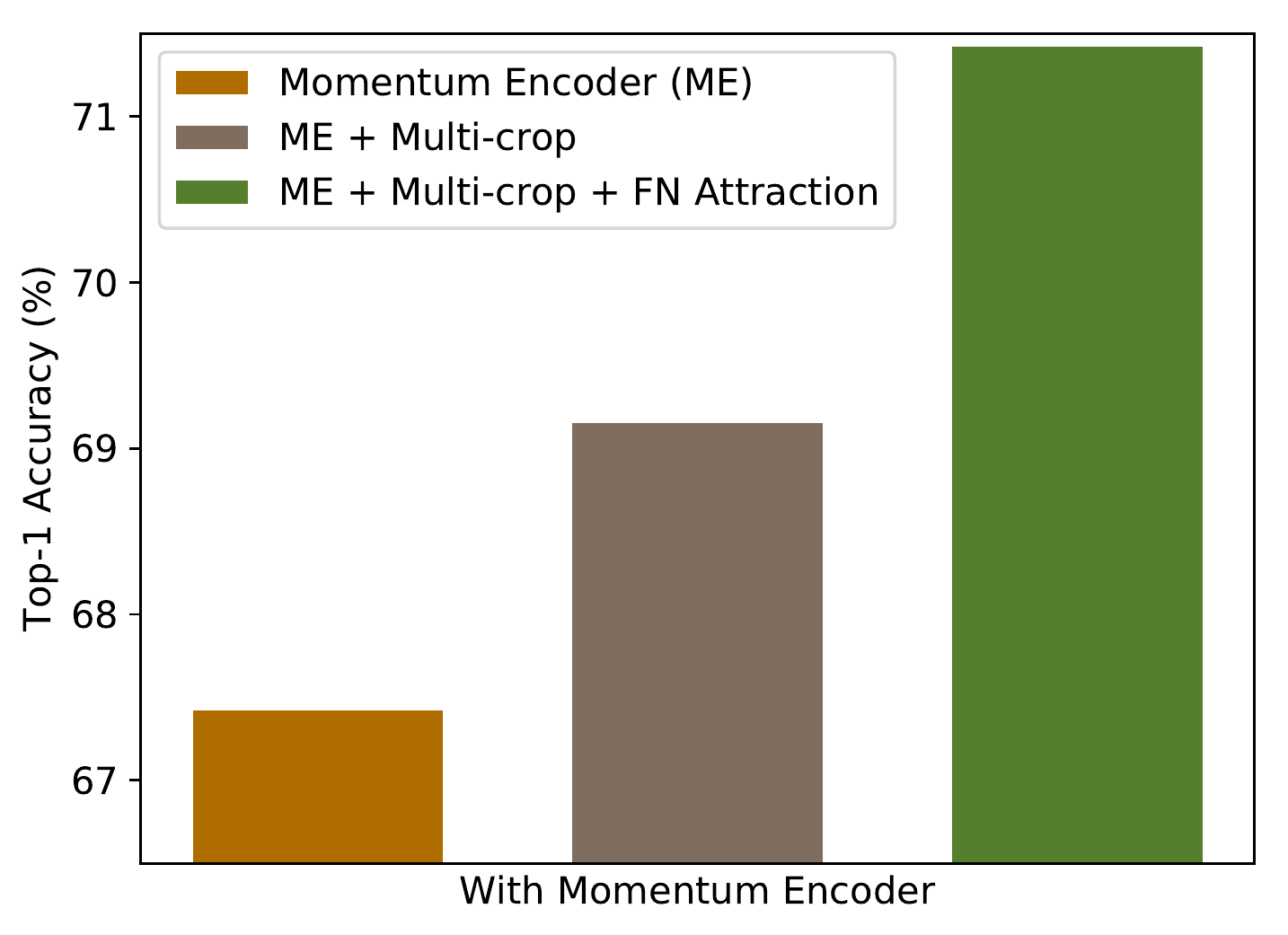}    
    \caption{False negative cancellation with (left) multi-crop and (right) momentum encoder.} \label{fig:multicrop_momentum}
    \vspace{-5pt}
\end{figure}
Thus far, we have identified false negatives in the current batch from a single encoder. However, other methods may store negative samples in a memory bank, or encoded by a momentum encoder. Here, we investigate the behaviors of the proposed method in these settings. Momentum contrast~\cite{he2019moco} employs two encoders, the main encoder and momentum encoder, and a memory bank where negative samples are stored alongside the samples from the current batch. This offers more options for finding false negatives, \emph{i.e.}, whether to use the support set from the main encoder or the momentum encoder, or whether to find negatives from samples in the current batch or all samples in memory. Figure~\ref{fig:baseline_vs_fn_encoder_fn_pool} (right) indicates that it is better to generate the support set using samples from the momentum encoder compared to the main encoder. Further, finding false negatives in the memory bank yields greater top-1 accuracy than drawing false negatives from the current batch.

\begin{table}[h]
    {\footnotesize
    \centering
    \setlength{\tabcolsep}{2pt}
    \begin{tabularx}{\linewidth}{Xccc}
    \toprule
    \multicolumn{2}{c}{Baseline} & + FN Cancel & Diff.\\
    \midrule
    SimCLR & 66.41 & 68.16 & \textcolor{ForestGreen}{+1.75} \\
    SimCLR + Multi-crop & 68.50 & 70.42 & \textcolor{ForestGreen}{+1.92} \\
    SimCLR + Multi-crop + Momentum & 69.16 & 71.43 & \textbf{\textcolor{ForestGreen}{+2.27}} \\
    \bottomrule
    \end{tabularx}}
    \vspace{3pt}
    \caption{Top-1 accuracy improvement of false negative cancellation for different baselines.}\label{tab:fn_improvement}
\end{table}
Figure~\ref{fig:multicrop_momentum} and Table~\ref{tab:fn_improvement} show that our method works across different configurations, with or without the presence of either momentum contrast or multi-crop. Notably, the performance gain from false negative cancellation in the presence of momentum contrast is even higher at $2.27\%$, a substantial boost in this large-scale ImageNet-1K setting. For context, this is larger than recent improvements in the state-of-the-art, such as BYOL over InfoMin ($1.3\%$)~\cite{grill2020bootstrap}, and MoCo over LocalAgg ($1.8\%$)~\cite{he2019moco}.


%
%
\begin{table}
    \footnotesize
    \centering
    \begin{tabularx}{\linewidth}{Xccc}
    \toprule
    Model & Epochs & Time (h) & Acc. (\%)\\
    \midrule
    SimCLR & \hphantom{0}100 & \hphantom{2}2.63 & 66.41 \\
    SimCLR & 1000 & 26.22 & 70.34 \\
    Improved Model & \hphantom{0}100 & \hphantom{2}7.50 & 70.42 \\
    \bottomrule
    \end{tabularx}
    \vspace{3pt}
    \caption{Computational efficiency and accuracy.} \label{tab:efficiency_final}
    \vspace{-10pt}
\end{table}

\subsubsection{With True Labels \& False Negatives Accuracy}
\begin{figure}[!b]
    \vspace{-5pt}
    \centering
    \includegraphics[width=0.51\linewidth,valign=t]{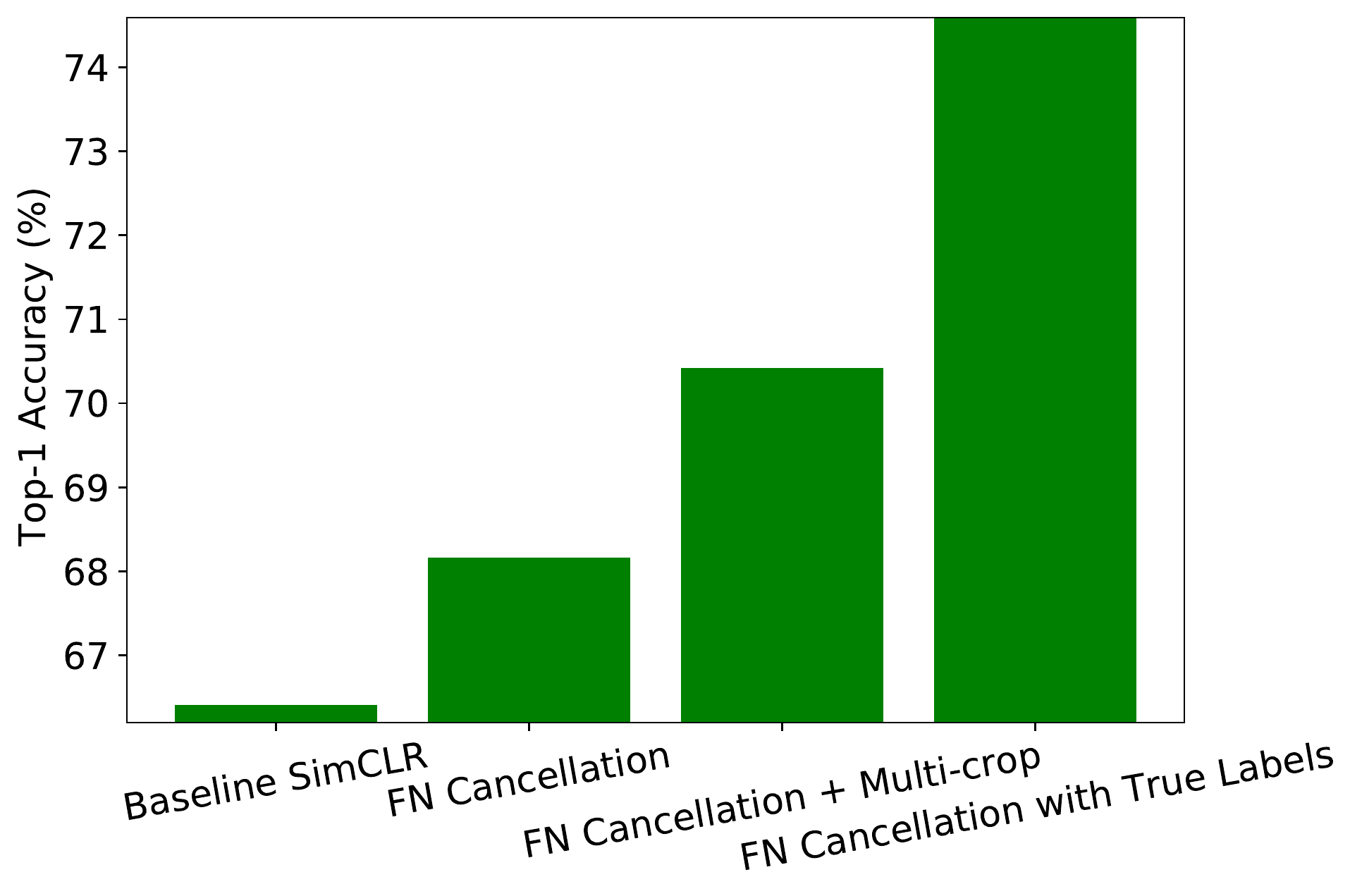}\hfil
    \includegraphics[width=0.45\linewidth,valign=t]{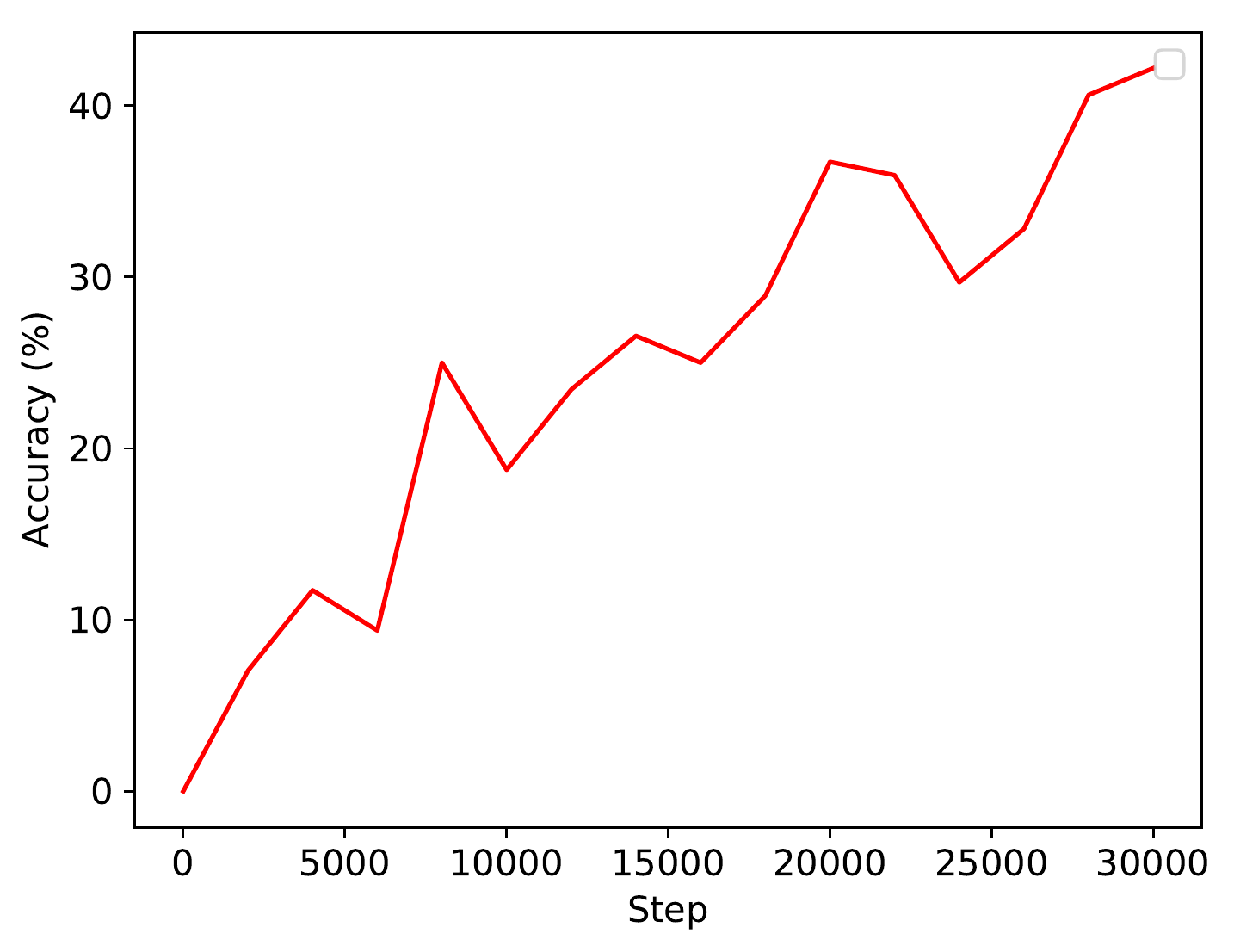}    
    \caption{A visualization of (left) top-1 accuracy with false negative cancellation using detected vs.\ ground-truth labels and (right) the accuracy of false negative detection.} \label{fig:fn_acc}
\end{figure}
We now consider the effectiveness of our false negative detection and cancellation strategy relative to the ideal setting in which we have access to ground-truth labels, which provides an upper-bound on performance. Figure~\ref{fig:fn_acc} presents the top-1 accuracy when using false negative cancellation combined with multi-crop data augmentation as well as the accuracy that results from false negative cancellation using ground-truth labels. As expected, cancelling false negatives helps substantially ($8.18\%$) when true labels are used. However, we close half the gap by just using multi-crop and the false negatives our method finds, increasing top-1 accuracy by $4.01\%$ over the SimCLR baseline.

To better understand the extent to which we are able to identify false negatives, Figure~\ref{fig:fn_acc} (right) plots the accuracy of the false negative detections over 100 epochs of pretraining. We see that the false negative detection accuracy steadily increases, reaching approximately $40\%$ accuracy by 100 epochs. Note that the false negatives accuracy is computed based on human-defined semantic labels, with 1000 categories in ImageNet. The chance of finding a false negative for an anchor at random is just $0.1\%$.

\begin{figure}[t]
    \vspace{-9pt}
    \hfill
    \begin{minipage}[t]{0.48\linewidth}
        \vspace{0pt}
        \includegraphics[width=1.0\linewidth]{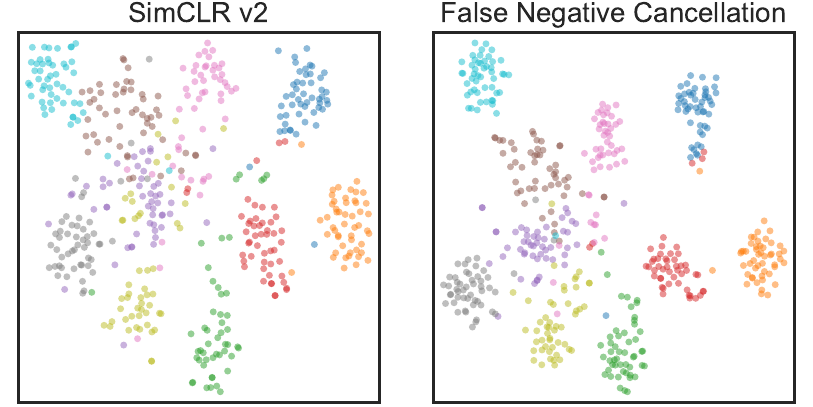}
    \end{minipage}
    \hfill\vline\hfill
    \begin{minipage}[t]{0.48\linewidth}
        \vspace{0pt}
        \includegraphics[width=1.0\linewidth]{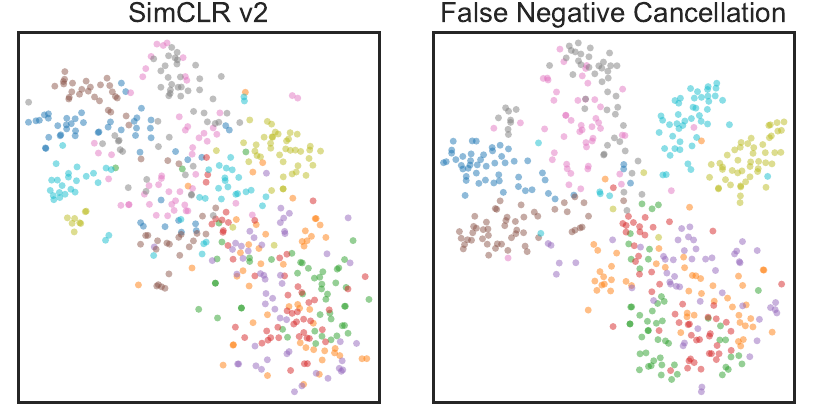}
    \end{minipage}
    \hfill
    \vspace{7pt}
    \caption{t-SNE visualizations of SimCLR and FNC for 10 random classes (left) and 10 dog classes (right) from ImageNet.}
    \label{fig:tsne}
    \vspace{-10pt}
\end{figure}

\vspace{-5pt}
\subsubsection{Computational Efficiency \& Visualization}

Table~\ref{tab:efficiency_final} compares the computational cost of our false negative cancellation strategy compared to the SimCLR baseline. As expected, we can see that for the same number of epochs, the process of detecting and incorporating false negatives incurs additional computational time. However, to achieve the same accuracy, SimCLR requires more than three times the amount of computation (26.22\,h and 1000 epochs) than our framework (7.50\,h and 100 epochs).

Figure~\ref{fig:tsne} shows that the embedding learned with false negative cancellation indeed demonstrates better class separation compared to baseline SimCLR.

\subsection{Comparison with State-of-the-Art}

\begin{table}[!b]
    \footnotesize
    \centering
    \begin{tabularx}{\linewidth}{Xcc}
    \toprule
    Method & top-1 & top-5\\
    \midrule
    Supervised & 76.5 &\\
    \hline
    \small{\textit{Representation Learning}} &&\\
    \hdashline
    \small{\textit{Contrastive learning}} &&\\
    MoCo v1~\cite{he2019moco} & 60.6 & --- \\ 
    PIRL~\cite{misra2019selfsupervised} & 63.6 & ---\\  
    PCL~\cite{li2020prototypical} & 65.9 & --- \\
    SimCLR v1~\cite{chen2020simple}& 69.3 & 89.0\\
    MoCo v2~\cite{chen2020mocov2}& 71.1 & ---\\
    SimCLR v2~\cite{chen2020big} & 71.7 & 90.4\\
    InfoMin~\cite{tian2020what} & 73.0 & 91.1\\
    FNC (ours) & \textit{\textbf{74.4}} & \textbf{91.8}\\
    \hdashline
    \small{\textit{Others}} &&\\
    BYOL~\cite{grill2020bootstrap} & 74.3& 91.6\\
    SwAV~\cite{caron2020unsupervised}& \textbf{75.3} & ---\\
    \hline
    \end{tabularx}
    \vspace{5pt}
    \caption{ImageNet linear evaluation.} \label{tab:imagenet_linear_eval}
\end{table}
%
%
%
%
\begin{table}[!b]
\footnotesize
\centering
\begin{tabularx}{\linewidth}{Xcccc}
\toprule
& \multicolumn{2}{c}{1\%} & \multicolumn{2}{c}{10\%}\\
Method & top-1 & top-5 & top-1 & top-5\\
\midrule
Supervised & 25.4 & 56.4 & 48.4 & 80.4\\
\hline
\footnotesize{\textit{Semi-supervised}}  &&&&\\
UDA~\cite{xie2019unsupervised}& --- & 68.8 & --- & 88.5\\
FixMatch~\cite{sohn2020fixmatch} & --- & 71.5 & --- & 89.1\\
\hline
\footnotesize{\textit{Representation Learning}} &&&&\\
\hdashline
\footnotesize{\textit{Contrastive learning}} &&&&\\
PIRL~\cite{misra2019selfsupervised} & 30.7 & 60.4 & 57.2 & 83.8\\  
PCL~\cite{li2020prototypical} & --- & --- & 75.6 & 86.2\\
SimCLR v1~\cite{chen2020simple} & 48.3 & 75.5 & 65.6 & 87.8\\
SimCLR v2~\cite{chen2020big}& 57.9 & 82.5 & 68.4 & 89.2\\
FNC (ours) & \textbf{63.7} & \textbf{85.3} & \textbf{71.1} & \textbf{90.2} \\
\hdashline
\footnotesize{\textit{Others}} &&&&\\
BYOL~\cite{grill2020bootstrap} & 53.2 & 78.4 & 68.8 & 89.0 \\
SwAV~\cite{caron2020unsupervised} & 53.9 & 78.5 & 70.2 & 89.9 \\

\bottomrule
\end{tabularx}
\vspace{5pt}
\caption{ImageNet semi-supervised evaluation.}
\label{tab:imagenet_semisup_eval}
\end{table}
\begin{table*}[!t]
\centering
\setlength{\tabcolsep}{4.5pt}
{\footnotesize
\begin{tabularx}{\linewidth}{lccccccccccccc}
\toprule
& Food & CIFAR10 & CIFAR100 & Birdsnap & SUN397 & Cars & Aircraft & VOC2007 & DTD & Pets & Caltech-101 & Flowers & Avg\\
\midrule
\textit{Linear eval}\\ 
SimCLR v1~\cite{chen2020simple}& 68.4 & 90.6 & 71.6 & 37.4 & 58.8 & 50.3 & 50.3 & 80.5 & 74.5 & 83.6 & 90.3 & 91.2 & 70.6\\
SimCLR v2~\cite{chen2020big}& 73.9 & 92.4 & 76.0 & 44.7 & 61.0 & 54.9 & 51.1 &81.2 &	\textbf{76.5} & 85.0 & 91.2 & 93.5 & 73.4\\
BYOL~\cite{grill2020bootstrap} & \textbf{75.3} & 91.3 & \textbf{78.4} & \textbf{57.2} & 62.2 & 67.8 & 60.6 & 82.5 & 75.5 & \textbf{90.4} & \textbf{94.2} & \textbf{96.1} & \textbf{77.6}\\
FNC (ours) & 74.4 & \textbf{93.0} & 76.8 & 54.0 & \textbf{63.2} & \textbf{68.8} & \textbf{61.3} & \textbf{83.0} & 76.3 & 89.0 & 93.5 & 94.9 & 77.3\\
\hline
\textit{Finetuned}\\
SimCLR v1~\cite{chen2020simple} & 88.2 &	97.7 & 85.9 &	75.9 & 63.5 & 91.3 & 88.1 & 84.1 & 73.2 & 89.2 & 92.1 & 97.0 & 85.5\\
SimCLR v2~\cite{chen2020big} & 88.2 &	97.5 & 86.0 &	74.9 &	\textbf{64.6} & 91.8 & 87.6 & 84.1 & 74.7 & 89.9 & 92.3 & 97.2 & 85.7\\
BYOL~\cite{grill2020bootstrap} & \textbf{88.5} & \textbf{97.8} & 86.1 & \textbf{76.3} & 63.7 & 91.6 & 88.1 & \textbf{85.4} & \textbf{76.2} & \textbf{91.7} & \textbf{93.8} & 97.0 & 86.3\\
FNC (ours) & 88.3 & 97.7 & \textbf{86.8} & \textbf{76.3} & 64.2 & \textbf{92.0} & \textbf{88.5} & 84.7 & 76.0 & 90.9 & 93.6 & \textbf{97.5} & \textbf{86.4}\\
\bottomrule
\end{tabularx}}
\vspace{5pt}
\caption{Transfer learning on classification task using ImageNet-pretrained ResNet models across 12 data sets.}
\label{tab:transfer_learning_class}
\end{table*}
%
%
\begin{table}
\centering
\begin{tabularx}{0.7\linewidth}{Xc}
\toprule
Method & AP50 \\ 
\midrule
Supervised & 81.3 \\
\hline
MoCo v2~\cite{chen2020mocov2} & 82.5 \\
SwAV~\cite{caron2020unsupervised} & 82.6 \\
FNC (ours) & \textbf{82.8}  \\
\bottomrule
\end{tabularx}
\vspace{7pt}
\caption{Transfer learning on Pascal VOC object detection.}
\label{tab:transfer_learning_det_seg}
\end{table}

We compare our improved model with false negative cancellation to other state-of-the-art methods on standard ImageNet evaluations and transfer to downstream tasks.

\paragraph{Pretraining Settings} We use similar configurations as SimCLR v2. Specifically, we use ResNet-50 as the base encoder, with a 3-layer MLP projection head. We use a 65k memory buffer for the momentum encoder, with a momentum of 0.999. Following \citet{tian2020what}, we use random crops, color distortion, Gaussian blur, and RandAugment~\cite{cubuk2019randaugment} 
for data augmentation. 
For false negative cancellation, we use the attraction strategy with max aggregation. The top-$k$ is set to 10 and a threshold of 0.7 is used for filtering the scores. The support size is 8, which is shared with multi-crop. We pretrain for 1000 epochs on 128 Cloud TPUs with a batch size of 4096. We use the LARS optimizer with a learning rate of 6.4, a cosine schedule for decaying the learning rate, and a weight decay of $1  \times 10^{-4}$.
\subsubsection{ImageNet Evaluation}

Our evaluation on ImageNet follows a protocol similar to that of SimCLR v2. Specifically, we conduct linear evaluation, in which a linear classifier is trained on top of the frozen pretrained features, and consider semi-supervised settings, in which we finetune the network with $1\%$ and $10\%$ labels available. We use a batch size of 1024 with a 0.16 learning rate, while weight decay is removed. We finetune for 90 epochs in linear evaluation, 60 epochs for $1\%$, and 30 epochs for $10\%$ labels in semi-supervised settings.

Table~\ref{tab:imagenet_linear_eval} presents the linear evaluation results. Our method achieves state-of-the-art performance for contrastive learning-based models with a top-1 accuracy of $74.4\%$, a $2.7\%$ improvement over SimCLR v2 and a $1.4\%$ boost from the previous best. Among all approaches, our method is second only to SwAV~\cite{caron2020unsupervised}, a clustering-based method, and reaches competitive results with BYOL~\cite{grill2020bootstrap}, a recent state-of-the-art method that only uses positive samples.

Table~\ref{tab:imagenet_semisup_eval} presents the results in the semi-supervised setting, in which our proposed model not only exceeds other contrastive learning-based methods, but also achieves the best performance among all models across all measures. The strong semi-supervised performance could be attributed to our improvement upon the baseline SimCLR v2, which is already quite good in this setting. Notably, in the $1\%$ labels case, our method significantly improves over the previous best to achieve $63.7\%$ top-1 accuracy (a $5.8\%$ absolute improvement).

\subsubsection{Transferring Features}

\paragraph{Image Classification} Following SimCLR v1, we perform the same evaluations on 12 classification datasets: Food~\cite{food101}, CIFAR10~\cite{cifar10-cifar100}, CIFAR100 ~\cite{cifar10-cifar100}, Birdsnap~\cite{birdsnap}, SUN397~\cite{sun397}, Cars~\cite{cars}, Aircraft~\cite{aircraft}, VOC2007~\cite{voc2007-voc2012}, DTD~\cite{DTD}, Pets~\cite{pets}, Caltech-101~\cite{caltech-101}, and Flowers~\cite{flowers}. We follow the same setup as in SimCLR v1. As Table~\ref{tab:transfer_learning_class} demonstrates, our approach achieves a significant improvement in performance among contrastive learning-based methods (i.e., SimCLR v1 and SimCLR v2) in both settings. In linear evaluation, our method outperforms SimCLR v1 on all 12 datasets and is better than SimCLR v2 on all but one dataset. In finetuning, the proposed model outperforms both SimCLR v1 and v2 on all but one dataset, and matches that of BYOL, with each being superior on about half of the datasets.
\vspace{-9pt}
\paragraph{Object Detection} To further evaluate the transferability of the learned embeddings, we finetune the model on PASCAL VOC object detection. We use similar settings as in MoCo~\cite{he2019moco}, where we finetune on the VOC \verb|trainval07+12| set using Faster R-CNN with a R50-C4 backbone, and evaluate on VOC \verb|test2007|. We train for 34K iterations with batch size 16. The learning rate is set to 0.02, which is reduced by a factor of 10 after 20K and 28K iterations.
As Table~\ref{tab:transfer_learning_det_seg} reveals, our proposed method outperforms both MoCo v2, SwAV, and the supervised baseline.

\section{Conclusion}
\label{sec:conclusion}
In this work, we address a fundamental problem in contrastive self-supervised learning that has not been adequately studied, identifying false negatives, and propose strategies to utilize this ability to improve contrastive learning frameworks. In addition to bringing novel insights to this topic through in-depth experimental analysis, our proposed method significantly boosts existing models, and sets new performance standards for contrastive self-supervised learning methods.

{\small
\bibliography{egbib}

\begin{thebibliography}{54}
\providecommand{\natexlab}[1]{#1}
\providecommand{\url}[1]{\texttt{#1}}
\expandafter\ifx\csname urlstyle\endcsname\relax
  \providecommand{\doi}[1]{doi: #1}\else
  \providecommand{\doi}{doi: \begingroup \urlstyle{rm}\Url}\fi

\bibitem[Asano et~al.(2020)Asano, Rupprecht, and Vedaldi]{asano2020self}
Y.~M. Asano, C.~Rupprecht, and A.~Vedaldi.
\newblock Self-labelling via simultaneous clustering and representation
  learning.
\newblock In \emph{Proc.\ Int'l Conf.\ on Learning Representations (ICLR)},
  2020.

\bibitem[Bachman et~al.(2019)Bachman, Hjelm, and
  Buchwalter]{bachman2019learning}
P.~Bachman, R.~D. Hjelm, and W.~Buchwalter.
\newblock Learning representations by maximizing mutual information across
  views.
\newblock In \emph{Advances in Neural Information Processing Systems
  (NeurIPS)}, pages 15535--15545, 2019.

\bibitem[Bautista et~al.(2016)Bautista, Sanakoyeu, Tikhoncheva, and
  Ommer]{cliquecnn2016}
M.~A. Bautista, A.~Sanakoyeu, E.~Tikhoncheva, and B.~Ommer.
\newblock Cliquecnn: Deep unsupervised exemplar learning.
\newblock In \emph{Advances in Neural Information Processing Systems
  (NeurIPS)}, pages 3846--3854, 2016.

\bibitem[Berg et~al.(2014)Berg, Liu, Lee, Alexander, Jacobs, and
  Belhumeur]{birdsnap}
T.~Berg, J.~Liu, S.~W. Lee, M.~L. Alexander, D.~W. Jacobs, and P.~N. Belhumeur.
\newblock Large-scale fine-grained visual categorization of birds.
\newblock In \emph{Proc.\ IEEE Conf.\ on Computer Vision and Pattern
  Recognition (CVPR)}, 2014.

\bibitem[Bossard et~al.(2014)Bossard, Guillaumin, and Gool]{food101}
L.~Bossard, M.~Guillaumin, and L.~V. Gool.
\newblock Food-101 {M}ining discriminative components with random forests.
\newblock In \emph{Proc.\ European Conf.\ on Computer Vision (ECCV)}, 2014.

\bibitem[Caron et~al.(2018)Caron, Bojanowski, Joulin, and Douze]{caron2018deep}
M.~Caron, P.~Bojanowski, A.~Joulin, and M.~Douze.
\newblock Deep clustering for unsupervised learning of visual features.
\newblock In \emph{Proc.\ European Conf.\ on Computer Vision (ECCV)}, 2018.

\bibitem[Caron et~al.(2020)Caron, Misra, Mairal, Goyal, Bojanowski, and
  Joulin]{caron2020unsupervised}
M.~Caron, I.~Misra, J.~Mairal, P.~Goyal, P.~Bojanowski, and A.~Joulin.
\newblock Unsupervised learning of visual features by contrasting cluster
  assignments.
\newblock In \emph{Advances in Neural Information Processing Systems
  (NeurIPS)}, 2020.

\bibitem[Chen et~al.(2020{\natexlab{a}})Chen, Radford, Child, Wu, Jun,
  Dhariwal, Luan, and Sutskever]{chen2020generative}
M.~Chen, A.~Radford, R.~Child, J.~Wu, H.~Jun, P.~Dhariwal, D.~Luan, and
  I.~Sutskever.
\newblock Generative pretraining from pixels.
\newblock In \emph{Proc.\ Int'l Conf.\ on Machine Learning (ICML)},
  2020{\natexlab{a}}.

\bibitem[Chen et~al.(2020{\natexlab{b}})Chen, Kornblith, Norouzi, and
  Hinton]{chen2020simple}
T.~Chen, S.~Kornblith, M.~Norouzi, and G.~Hinton.
\newblock A simple framework for contrastive learning of visual
  representations.
\newblock In \emph{Proc.\ Int'l Conf.\ on Machine Learning (ICML)},
  2020{\natexlab{b}}.

\bibitem[Chen et~al.(2020{\natexlab{c}})Chen, Kornblith, Swersky, Norouzi, and
  Hinton]{chen2020big}
T.~Chen, S.~Kornblith, K.~Swersky, M.~Norouzi, and G.~Hinton.
\newblock Big self-supervised models are strong semi-supervised learners.
\newblock In \emph{Advances in Neural Information Processing Systems
  (NeurIPS)}, 2020{\natexlab{c}}.

\bibitem[Chen et~al.(2020{\natexlab{d}})Chen, Fan, Girshick, and
  He]{chen2020mocov2}
X.~Chen, H.~Fan, R.~Girshick, and K.~He.
\newblock Improved baselines with momentum contrastive learning.
\newblock \emph{arXiv preprint arXiv:2003.04297}, 2020{\natexlab{d}}.

\bibitem[Chuang et~al.(2020)Chuang, Robinson, Yen-Chen, Torralba, and
  Jegelka]{chuang2020debiased}
C.-Y. Chuang, J.~Robinson, L.~Yen-Chen, A.~Torralba, and S.~Jegelka.
\newblock Debiased contrastive learning.
\newblock In \emph{Advances in Neural Information Processing Systems
  (NeurIPS)}, 2020.

\bibitem[Cimpoi et~al.(2014)Cimpoi, Maji, Kokkinos, Mohamed, and Vedaldi]{DTD}
M.~Cimpoi, S.~Maji, I.~Kokkinos, S.~Mohamed, and A.~Vedaldi.
\newblock Describing textures in the wild.
\newblock In \emph{Proc.\ IEEE Conf.\ on Computer Vision and Pattern
  Recognition (CVPR)}, 2014.

\bibitem[Cubuk et~al.(2020)Cubuk, Zoph, Shlens, and Le]{cubuk2019randaugment}
E.~D. Cubuk, B.~Zoph, J.~Shlens, and Q.~V. Le.
\newblock Randaugment: Practical data augmentation with no separate search.
\newblock In \emph{Proc.\ IEEE Conf.\ on Computer Vision and Pattern
  Recognition (CVPR)}, 2020.

\bibitem[Doersch et~al.(2015)Doersch, Gupta, and
  Efros]{doersch2016unsupervised}
C.~Doersch, A.~Gupta, and A.~A. Efros.
\newblock Unsupervised visual representation learning by context prediction.
\newblock In \emph{Proc.\ Int'l.\ Conf.\ on Computer Vision (ICCV)}, 2015.

\bibitem[Donahue et~al.(2014)Donahue, Jia, Vinyals, Hoffman, Zhang, Tzeng, and
  Darrell]{pmlr-v32-donahue14}
J.~Donahue, Y.~Jia, O.~Vinyals, J.~Hoffman, N.~Zhang, E.~Tzeng, and T.~Darrell.
\newblock {DeCAF}: {A} deep convolutional activation feature for generic visual
  recognition.
\newblock In \emph{Proc.\ Int'l Conf.\ on Machine Learning (ICML)}, pages
  647--655, 2014.

\bibitem[Everingham et~al.(2010)Everingham, Gool, Williams, Winn, and
  Zisserman]{voc2007-voc2012}
M.~Everingham, L.~V. Gool, C.~K. Williams, J.~Winn, and A.~Zisserman.
\newblock The {P}ascal visual object classes ({VOC}) challenge.
\newblock \emph{Int'l J.\ on Computer Vision}, 88\penalty0 (2):\penalty0
  303--338, 2010.

\bibitem[Fei-Fei et~al.(2004)Fei-Fei, Fergus, and Perona]{caltech-101}
L.~Fei-Fei, R.~Fergus, and P.~Perona.
\newblock Learning generative visual models from few training examples: An
  incremental {B}ayesian approach tested on 101 object categories.
\newblock In \emph{Proceedings of the CVPR Workshop on Generative-Model Based
  Vision}, 2004.

\bibitem[Geirhos et~al.(2020)Geirhos, Jacobsen, Michaelis, Zemel, Brendel,
  Bethge, and Wichmann]{geirhos2020shortcut}
R.~Geirhos, J.-H. Jacobsen, C.~Michaelis, R.~Zemel, W.~Brendel, M.~Bethge, and
  F.~A. Wichmann.
\newblock Shortcut learning in deep neural networks.
\newblock \emph{arXiv preprint arXiv:2004.07780}, 2020.

\bibitem[Gidaris et~al.(2018)Gidaris, Singh, and
  Komodakis]{gidaris2018unsupervised}
S.~Gidaris, P.~Singh, and N.~Komodakis.
\newblock Unsupervised representation learning by predicting image rotations.
\newblock In \emph{Proc.\ Int'l Conf.\ on Learning Representations (ICLR)},
  2018.

\bibitem[Ginsburg et~al.(2018)Ginsburg, Gitman, and You]{ginsburg2018large}
B.~Ginsburg, I.~Gitman, and Y.~You.
\newblock Large batch training of convolutional networks with layer-wise
  adaptive rate scaling.
\newblock In \emph{Proc.\ Int'l Conf.\ on Learning Representations (ICLR)},
  2018.

\bibitem[Girshick et~al.(2014)Girshick, Donahue, Darrell, and Malik]{6909475}
R.~Girshick, J.~Donahue, T.~Darrell, and J.~Malik.
\newblock Rich feature hierarchies for accurate object detection and semantic
  segmentation.
\newblock In \emph{Proc.\ IEEE Conf.\ on Computer Vision and Pattern
  Recognition (CVPR)}, pages 580--587, 2014.

\bibitem[Grill et~al.(2020)Grill, Strub, Altché, Tallec, Richemond,
  Buchatskaya, Doersch, Pires, Guo, Azar, Piot, Kavukcuoglu, Munos, and
  Valko]{grill2020bootstrap}
J.-B. Grill, F.~Strub, F.~Altché, C.~Tallec, P.~H. Richemond, E.~Buchatskaya,
  C.~Doersch, B.~A. Pires, Z.~D. Guo, M.~G. Azar, B.~Piot, K.~Kavukcuoglu,
  R.~Munos, and M.~Valko.
\newblock Bootstrap your own latent: A new approach to self-supervised
  learning.
\newblock \emph{arXiv preprint arXiv:2006.07733}, 2020.

\bibitem[He et~al.(2019)He, Fan, Wu, Xie, and Girshick]{he2019moco}
K.~He, H.~Fan, Y.~Wu, S.~Xie, and R.~Girshick.
\newblock Momentum contrast for unsupervised visual representation learning.
\newblock In \emph{Proc.\ IEEE Conf.\ on Computer Vision and Pattern
  Recognition (CVPR)}, 2019.

\bibitem[Henaff et~al.(2020)Henaff, Srinivas, Fauw, Razavi, Doersch, Eslami,
  and van~den Oord]{henaff2020dataefficient}
O.~J. Henaff, A.~Srinivas, J.~D. Fauw, A.~Razavi, C.~Doersch, S.~M.~A. Eslami,
  and A.~van~den Oord.
\newblock Data-efficient image recognition with contrastive predictive coding.
\newblock In \emph{Proc.\ Int'l Conf.\ on Learning Representations (ICLR)},
  2020.

\bibitem[Hjelm et~al.(2019)Hjelm, Fedorov, Lavoie-Marchildon, Grewal, Bachman,
  Trischler, and Bengio]{hjelm2018learning}
R.~D. Hjelm, A.~Fedorov, S.~Lavoie-Marchildon, K.~Grewal, P.~Bachman,
  A.~Trischler, and Y.~Bengio.
\newblock Learning deep representations by mutual information estimation and
  maximization.
\newblock In \emph{Proc.\ Int'l Conf.\ on Learning Representations (ICLR)},
  2019.

\bibitem[Kalantidis et~al.(2020)Kalantidis, Sariyildiz, Pion, Weinzaepfel, and
  Larlus]{kalantidis2020hard}
Y.~Kalantidis, M.~B. Sariyildiz, N.~Pion, P.~Weinzaepfel, and D.~Larlus.
\newblock Hard negative mixing for contrastive learning.
\newblock In \emph{Advances in Neural Information Processing Systems
  (NeurIPS)}, 2020.

\bibitem[Khosla et~al.(2020)Khosla, Teterwak, Wang, Sarna, Tian, Isola,
  Maschinot, Liu, and Krishnan]{khosla2020supervised}
P.~Khosla, P.~Teterwak, C.~Wang, A.~Sarna, Y.~Tian, P.~Isola, A.~Maschinot,
  C.~Liu, and D.~Krishnan.
\newblock Supervised contrastive learning.
\newblock \emph{arXiv preprint arXiv:2004.11362}, 2020.

\bibitem[Kim and Walter(2017)]{kim17}
D.-K. Kim and M.~R. Walter.
\newblock Satellite image-based localization via learned embeddings.
\newblock In \emph{Proc.\ IEEE Int'l Conf.\ on Robotics and Automation (ICRA)},
  Singapore, May 2017.

\bibitem[Krause et~al.(2013)Krause, Dengand, and Fei-Fei]{cars}
J.~Krause, M.~S.~J. Dengand, and L.~Fei-Fei.
\newblock {3D} object representations for fine-grained categorization.
\newblock In \emph{Proc.\ Int'l.\ Conf.\ on Computer Vision (ICCV)}, 2013.

\bibitem[Krizhevsky(2009)]{cifar10-cifar100}
A.~Krizhevsky.
\newblock Learning multiple layers of features from tiny images.
\newblock \emph{Technical report, University of Toronto.}, 2009.

\bibitem[{Larsson} et~al.(2017){Larsson}, {Maire}, and
  {Shakhnarovich}]{8099579}
G.~{Larsson}, M.~{Maire}, and G.~{Shakhnarovich}.
\newblock Colorization as a proxy task for visual understanding.
\newblock In \emph{2017 IEEE Conference on Computer Vision and Pattern
  Recognition (CVPR)}, pages 840--849, 2017.

\bibitem[Li et~al.(2020)Li, Zhou, Xiong, Socher, and Hoi]{li2020prototypical}
J.~Li, P.~Zhou, C.~Xiong, R.~Socher, and S.~C.~H. Hoi.
\newblock Prototypical contrastive learning of unsupervised representations.
\newblock \emph{arXiv preprint arXiv:2005.04966}, 2020.

\bibitem[Maji et~al.(2013)Maji, Rahtu, Kannala, Blaschko, and
  Vedaldi]{aircraft}
S.~Maji, E.~Rahtu, J.~Kannala, M.~B. Blaschko, and A.~Vedaldi.
\newblock Fine-grained visual classification of aircraft.
\newblock \emph{arXiv preprint arXiv:1306.5151}, 2013.

\bibitem[Misra and van~der Maaten(2020)]{misra2019selfsupervised}
I.~Misra and L.~van~der Maaten.
\newblock Self-supervised learning of pretext-invariant representations.
\newblock In \emph{Proc.\ IEEE Conf.\ on Computer Vision and Pattern
  Recognition (CVPR)}, 2020.

\bibitem[Nilsback and Zisserman(2008)]{flowers}
M.-E. Nilsback and A.~Zisserman.
\newblock Automated flower classification over a large number of classes.
\newblock In \emph{Proc.\ of the Indian Conf.\ on Computer Vision, Graphics and
  Image Processing}, 2008.

\bibitem[Noroozi and Favaro(2016)]{noroozi2017unsupervised}
M.~Noroozi and P.~Favaro.
\newblock Unsupervised learning of visual representations by solving jigsaw
  puzzles.
\newblock In \emph{Proc.\ European Conf.\ on Computer Vision (ECCV)}, 2016.

\bibitem[Parkhi et~al.(2012)Parkhi, Vedaldi, Zisserman, and Jawahar]{pets}
O.~M. Parkhi, A.~Vedaldi, A.~Zisserman, and C.~V. Jawahar.
\newblock Cats and dogs.
\newblock In \emph{Proc.\ IEEE Conf.\ on Computer Vision and Pattern
  Recognition (CVPR)}, 2012.

\bibitem[Pathak et~al.(2016)Pathak, Krahenbuhl, Donahue, Darrell, and
  Efros]{pathak2016context}
D.~Pathak, P.~Krahenbuhl, J.~Donahue, T.~Darrell, and A.~A. Efros.
\newblock Context encoders: Feature learning by inpainting.
\newblock In \emph{Proc.\ IEEE Conf.\ on Computer Vision and Pattern
  Recognition (CVPR)}, 2016.

\bibitem[Robinson et~al.(2020)Robinson, Chuang, Sra, and
  Jegelka]{robinson2020contrastive}
J.~Robinson, C.-Y. Chuang, S.~Sra, and S.~Jegelka.
\newblock Contrastive learning with hard negative samples.
\newblock \emph{arXiv preprint arXiv:2010.04592}, 2020.

\bibitem[Russakovsky et~al.(2015)Russakovsky, Deng, Su, Krause, Satheesh, Ma,
  Huang, Karpathy, Khosla, Bernstein, Berg, and Fei-Fei]{imagenet}
O.~Russakovsky, J.~Deng, H.~Su, J.~Krause, S.~Satheesh, S.~Ma, Z.~Huang,
  A.~Karpathy, A.~Khosla, M.~Bernstein, A.~C. Berg, and L.~Fei-Fei.
\newblock {ImageNet} large scale visual recognition challenge.
\newblock \emph{International Journal of Computer Vision}, 115(3):211–252,
  2015.

\bibitem[Sohn et~al.(2020)Sohn, Berthelot, Li, Zhang, Carlini, Cubuk, Kurakin,
  Zhang, and Raffel]{sohn2020fixmatch}
K.~Sohn, D.~Berthelot, C.-L. Li, Z.~Zhang, N.~Carlini, E.~D. Cubuk, A.~Kurakin,
  H.~Zhang, and C.~Raffel.
\newblock Fixmatch: Simplifying semi-supervised learning with consistency and
  confidence.
\newblock \emph{arXiv preprint arXiv:2001.07685}, 2020.

\bibitem[Tian et~al.(2020{\natexlab{a}})Tian, Krishnan, and
  Isola]{tian2020contrastive}
Y.~Tian, D.~Krishnan, and P.~Isola.
\newblock Contrastive multiview coding.
\newblock In \emph{Proc.\ European Conf.\ on Computer Vision (ECCV)},
  2020{\natexlab{a}}.

\bibitem[Tian et~al.(2020{\natexlab{b}})Tian, Sun, Poole, Krishnan, Schmid, and
  Isola]{tian2020what}
Y.~Tian, C.~Sun, B.~Poole, D.~Krishnan, C.~Schmid, and P.~Isola.
\newblock What makes for good views for contrastive learning.
\newblock In \emph{Proc.\ European Conf.\ on Computer Vision (ECCV)},
  2020{\natexlab{b}}.

\bibitem[van~den Oord et~al.(2019)van~den Oord, Li, and
  Vinyals]{oord2019representation}
A.~van~den Oord, Y.~Li, and O.~Vinyals.
\newblock Representation learning with contrastive predictive coding.
\newblock In \emph{Proc.\ Int'l Conf.\ on Machine Learning (ICML)}, 2019.

\bibitem[Vincent et~al.(2008)Vincent, Larochelle, Bengio, and
  Manzagol]{pascal2018extracting}
P.~Vincent, H.~Larochelle, Y.~Bengio, and P.-A. Manzagol.
\newblock Extracting and composing robust features with denoising autoencoders.
\newblock In \emph{Proc.\ Int'l Conf.\ on Machine Learning (ICML)}, 2008.

\bibitem[Wang et~al.(2020)Wang, Liu, Guo, and Fuchun]{wang2020advances}
F.~Wang, H.~Liu, D.~Guo, and S.~Fuchun.
\newblock Unsupervised representation learning by invariance propagation.
\newblock In H.~Larochelle, M.~Ranzato, R.~Hadsell, M.~F. Balcan, and H.~Lin,
  editors, \emph{Advances in Neural Information Processing Systems (NeurIPS)},
  volume~33, pages 3510--3520, 2020.

\bibitem[Wu et~al.(2021)Wu, Mosse, Zhuang, Yamins, and
  Goodman]{wu2021conditional}
M.~Wu, M.~Mosse, C.~Zhuang, D.~Yamins, and N.~Goodman.
\newblock Conditional negative sampling for contrastive learning of visual
  representations.
\newblock In \emph{Proc.\ Int'l Conf.\ on Learning Representations (ICLR)},
  2021.

\bibitem[Xiao et~al.(2010)Xiao, Hays, Ehinger, Oliva, and Torralba]{sun397}
J.~Xiao, J.~Hays, K.~A. Ehinger, A.~Oliva, and A.~Torralba.
\newblock {SUN} {D}atabase: {L}arge-scale scene recognition from abbey to zoo.
\newblock In \emph{Proc.\ IEEE Conf.\ on Computer Vision and Pattern
  Recognition (CVPR)}, 2010.

\bibitem[Xie et~al.(2016)Xie, Girshick, and Farhadi]{10.5555/3045390.3045442}
J.~Xie, R.~Girshick, and A.~Farhadi.
\newblock Unsupervised deep embedding for clustering analysis.
\newblock In \emph{Proc.\ Int'l Conf.\ on Machine Learning (ICML)}, page
  478–487, 2016.

\bibitem[Xie et~al.(2019)Xie, Dai, Hovy, Luong, and Le]{xie2019unsupervised}
Q.~Xie, Z.~Dai, E.~Hovy, M.-T. Luong, and Q.~V. Le.
\newblock Unsupervised data augmentation for consistency training.
\newblock \emph{arXiv preprint arXiv:1904.12848}, 2019.

\bibitem[Zeiler and Fergus(2014)]{10.1007/978-3-319-10590-1_53}
M.~D. Zeiler and R.~Fergus.
\newblock Visualizing and understanding convolutional networks.
\newblock In \emph{Proc.\ European Conf.\ on Computer Vision (ECCV)}, pages
  818--833, 2014.

\bibitem[Zhang et~al.(2017)Zhang, Isola, and Efros]{8099559}
R.~Zhang, P.~Isola, and A.~A. Efros.
\newblock Split-brain autoencoders: Unsupervised learning by cross-channel
  prediction.
\newblock In \emph{Proc.\ IEEE Conf.\ on Computer Vision and Pattern
  Recognition (CVPR)}, pages 645--654, 2017.

\bibitem[Zheng et~al.(2021)Zheng, Chen, Yao, Yang, Li, Zhang, Zhang, Tsang,
  Zhou, and Zhou]{zheng2021contrastive}
H.~Zheng, X.~Chen, J.~Yao, H.~Yang, C.~Li, Y.~Zhang, H.~Zhang, I.~Tsang,
  J.~Zhou, and M.~Zhou.
\newblock Contrastive attraction and contrastive repulsion for representation
  learning.
\newblock \emph{arXiv preprint arXiv:2105.03746}, 2021.

\end{thebibliography}
}

\end{document}